%% file: neurips_2026.tex
\documentclass{article}

    \PassOptionsToPackage{numbers, compress}{natbib}
 \usepackage[preprint]{neurips_2026}

\usepackage[utf8]{inputenc} 
\usepackage[T1]{fontenc}    
\usepackage[colorlinks]{hyperref}
\usepackage{url}            
\usepackage{booktabs}       
\usepackage{amsfonts}       
\usepackage{nicefrac}       
\usepackage{microtype}      
\usepackage{xcolor}         
\usepackage[dvipsnames]{xcolor}         
\usepackage{xspace}
\usepackage{color}

\usepackage{booktabs}
\usepackage{tikz}
\usepackage{pgfplots}
\pgfplotsset{compat=1.18} 
\usepgfplotslibrary{groupplots}
\usepackage{multirow}

\usetikzlibrary{decorations.pathmorphing, calc}

\usepackage{graphicx}   
\usepackage{booktabs}   
\usepackage{amssymb}    
\usepackage{amsmath}    

\usepackage{bbding}
\usepackage{soul}
\usepackage{tabularray}

\newcommand{\parhead}[1]{\noindent\textbf{#1}}
\newcommand{\datasetname}{\texttt{AnyHand}\xspace}
\newcommand{\datasetnamedpx}{\texttt{\datasetname-Single}\xspace}
\newcommand{\datasetnamegxl}{\texttt{\datasetname-Interact}\xspace}
\newcommand{\rgbdmodel}{\texttt{{\datasetname}Net-D}\xspace}

\newcommand{\eg}{e.g.\xspace}

\newcommand{\etal}{et al.\xspace}
\newcommand{\vs}{vs.\xspace}

\title{AnyHand: A Large-Scale Synthetic Dataset for RGB(-D) Hand Pose Estimation}

\author{%
\textbf{Chen Si\textsuperscript{1}\thanks{Equal contribution.} \quad
Yulin Liu\textsuperscript{1}\footnotemark[1] \quad
Bo Ai\textsuperscript{1} \quad
Jianwen Xie\textsuperscript{2}}
\\
\textbf{Rolandos Alexandros Potamias\textsuperscript{3} \quad
Chuanxia Zheng\textsuperscript{4} \quad
Hao Su\textsuperscript{1}}
\\[4pt]
\textsuperscript{1}University of California, San Diego
\quad
\textsuperscript{2}Lambda, Inc
\\
\textsuperscript{3}Imperial College London
\quad
\textsuperscript{4}Nanyang Technological University
\\[4pt]
\url{https://chen-si-cs.github.io/projects/AnyHand}
}

\begin{document}

\maketitle

\input{sec/0_abstract.tex}
\input{figs/teaser.tex}
\input{sec/1_intro.tex}

\input{sec/2_related.tex}

\input{sec/3_data.tex}
\input{sec/4_RGB_exp}
\input{sec/6_conclusion.tex}

\medskip

\bibliographystyle{plainnat}
\bibliography{chuanxia_general,chuanxia_specific,anyhand_specific}

\newpage
\appendix

\section*{Appendix}

\input{sup_sec/rgb_exp}
\input{sec/5_RGBD_model}

\input{sup_sec/dataset}
\input{sup_sec/related_extended}

\input{sup_sec/computation}
\input{sup_sec/limitation}
\input{sup_sec/benchmark}





\end{document}

%% file: sec/0_abstract.tex
\begin{abstract}

We present {\datasetname}, a large-scale synthetic dataset designed to advance the state of the art in 3D hand pose estimation.
While recent 
works with 
foundation approaches have shown that 
scaling training data markedly improves hand pose estimation, 
existing real-world datasets are limited in coverage,
and prior synthetic datasets rarely provide occlusions, arm details, and aligned depth together at scale.
To address this bottleneck,
we propose \datasetname contains 2.5M single-hand and 4.1M hand-object interaction RGB-D images, with rich geometric annotations.
We show that extending the original training data recipes of existing RGB baselines with \datasetname yields significant gains on multiple benchmarks (FreiHAND and HO-3D), even when keeping the architectures and training schemes fixed.
More impressively, the resulting models generalize better
to the out-of-domain HO-Cap dataset at zero-shot testing.
Together with extensive ablations on the scale and composition of the training data setups, these results suggest that training data diversity and quality are as critical as scale for advancing hand pose estimation.
We further examine the utility of {\datasetname}'s aligned depth maps in the appendix, showing that scaling RGB-D supervision with {\datasetname} allows a lightweight depth-fusion variant of existing RGB baselines to outperform prior RGB-D methods.  

\end{abstract}

%% file: figs/teaser.tex
\begin{figure*}[t]
    \centering

    \includegraphics[width=\linewidth]{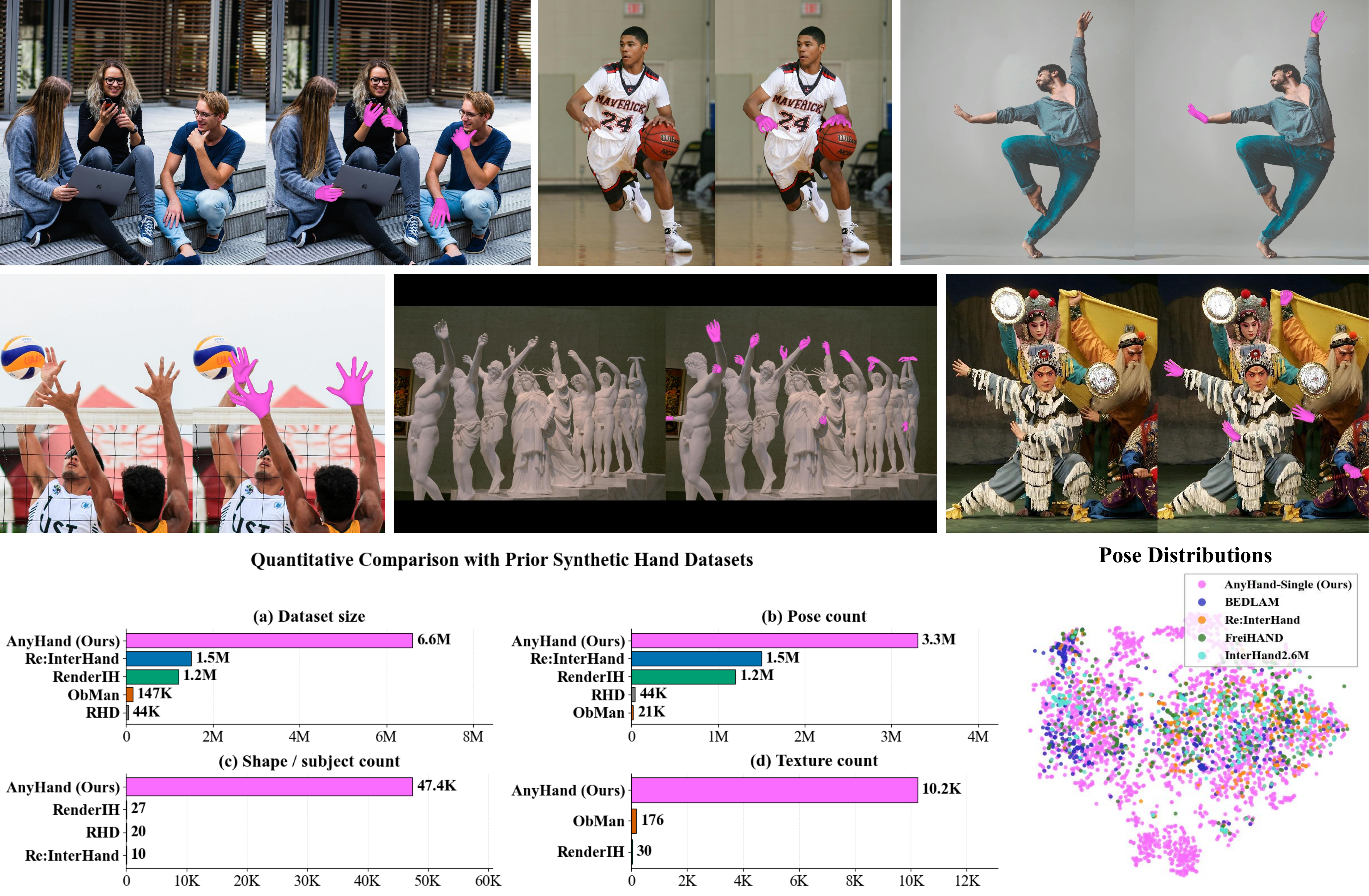}
    
    \caption{
    We propose {\datasetname} as a large-scale synthetic RGB-D dataset that substantially expands coverage of hand poses, hand-object interactions, textures, and viewpoint variations.
    When added to training data for models such as HaMeR~\cite{pavlakos2024hamer} and WiLoR~\cite{potamias2024wilor}, it yields consistent improvements across multiple benchmarks.
    Predicted hand meshes on in-the-wild images from WiLoR trained with \datasetname in addition are shown in pink.
    \textbf{Bottom-left}: Quantitative comparison with representative prior synthetic hand datasets across four axes: dataset size, unique pose count, shape/subject count, and texture count. AnyHand surpasses all prior datasets on every axis by a large margin.
    \textbf{Bottom-right}: t-SNE visualization of hand pose distributions across datasets. {\datasetnamedpx} densely covers the pose distributions of commonly used real and synthetic hand datasets, while also extending to additional low-density and edge-case regions that are less represented in prior data.
    \vspace{-0.35in}
    }
    \label{fig:teaser}
\end{figure*}

%% file: sec/1_intro.tex
\vspace{-0.1in}
\section{Introduction}%
\label{sec:intro}
\vspace{-0.1in}

Our daily interactions with the physical world are largely mediated by our hands in a 3D space,
a capacity widely regarded as \emph{a genesis of human intelligence}.
Grasping tools, taking a coffee, or typing on a keyboard are all examples of the remarkable fine-grained dexterity of human hands in manipulating objects of diverse shapes, sizes, and affordances.
Equipping machines with similar capabilities is crucial in VR / AR \cite{Apple2024VisionPro,Meta2023Quest3} and robotics \cite{cheng2024open,ding2024bunny,li2025maniptrans,mandi2025dexmachina,pan2025spider,shi2025learning,yang2025egovla},
where accurate hand pose estimation from visual observations is necessary for natural interaction with both the virtual and physical worlds.

In this work,
we consider the problem of \emph{3D hand pose estimation},
which aims to 
build models that can robustly estimate 3D hand pose from RGB
inputs across diverse real-world scenarios.
Several recent advances have been made in this direction,
driven by large-capacity transformer-based pipelines that regress parametric hand representations such as MANO~\cite{romero2017mano} from a single image.
Recent contributions such as HaMeR~\cite{pavlakos2024hamer}, Hamba~\cite{dong2024hamba}, and WiLoR~\cite{potamias2024wilor} demonstrate that relatively simple architectures perform well when trained on diverse, large-scale data.
However, scaling such data coverage needed for \emph{foundational} training remains difficult in practice.
For instance,
GigaHands~\cite{fu2025gigahands} provides sequential hand-object interaction annotations, but as a real-captured dataset, its diversity is constrained by the capture setup/collection scale (\eg viewing perspectives, subjects, and objects), and its 3D annotations might include noisy ones due to limitations of the annotation/reconstruction pipeline, especially under heavy occlusion.

Recent large-scale synthetic 3D corpora such as Objaverse(-XL)~\cite{deitke2023objaverse,deitke2023objaversexl} demonstrate a practical alternative: scaling synthetic data can measurably improve downstream 3D models on various of tasks~\cite{wen2024foundationpose,sam3dteam2025sam3d3dfyimages,zhang2024clay,xiang2024structured,hunyuan3d22025tencent,li2024puppet,li2025dso,wu2025amodal3r,jiang2026mesh4d}. 
Inspired by this data-scaling paradigm,
we introduce \datasetname, a large-scale RGB-D synthetic dataset that consists of 
hand-only and hand-object interaction scenes with realistic textures and rich annotations (Fig.~\ref{fig:teaser}).
The resulting dataset provides \emph{guaranteed} ground-truth labels,
as the synthetic hands' poses are ``perfectly labeled'' by construction, avoiding annotation noise common in real datasets.
It substantially expands visual diversity by randomizing camera viewpoints, backgrounds, and illumination, and by increasing subject-level variation in hand texture, skin tone, and hand shape beyond what is feasible with a limited pool of real participants. 
As a concrete example, we augment the attached-arm context with diverse realistic forearm appearances, including both bare skin and clothing such as short- and long-sleeves, which helps mitigate overfitting to clean capture conditions.
\textbf{To support reproducibility and future research, we will publicly release the full dataset, generation pipeline, and model checkpoints.}

We then design a mixed-training recipe that combines our created large-scale synthetic dataset \datasetname with existing datasets and validate the state-of-the-art models across multiple standard benchmarks.
Without bells and whistles,
that is, using the \emph{de facto} architecture and exactly the same training protocol as HaMeR~\cite{pavlakos2024hamer} and WiLoR~\cite{potamias2024wilor},
our retrained models surpass all previous state-of-the-art methods, suggesting that scaling training data can be a stronger lever than architecture design. Ablations on data scaling, real-synthetic mixing, and pipeline components further show that both the scale and the quality of the synthetic data contribute to these gains.


In summary, our main contributions are as follows:
\begin{itemize}
    \item We introduce \datasetname, a large-scale RGB-D synthetic hand dataset and a released generation pipeline that includes realistic single-hand and hand-object interaction scenes with aligned depth and arm context.
    \vspace{-0.025in}
    \item We show the benefits of our synthetic data through extensive experiments on standard benchmarks, demonstrating consistent improvements in both in-domain accuracy and out-of-domain generalization across diverse and challenging conditions.
    \vspace{-0.025in}
    \item We provide ablations on data scaling, real-synthetic mixing, and dataset components, offering practical insights into effective synthetic data design for hand pose estimation.
\end{itemize}

%% file: sec/2_related.tex
\vspace{-0.1in}
\section{Related works}%
\label{sec:related}
\vspace{-0.1in}

\parhead{3D hand pose estimation.}
Early 3D hand pose estimation approaches often relied on depth-based tracking, using geometric cues from depth and articulated alignment for real-time recovery~\cite{qian2014realtime, tagliasacchi2015robust}.
To move beyond depth sensors, Boukhayma \etal~\cite{boukhayma20193d} introduced the first fully learnable pipeline that regresses MANO~\cite{romero2017mano} parameters from RGB images.
Subsequent RGB-based methods improved reconstruction with stronger 2D supervision, refinement modules, mesh/vertex regression, and mesh convolutions~\cite{zhang2019end,baek2019pushing,kulon2019single,kulon2020weakly}.
Other works improve robustness under occlusion and motion blur~\cite{park2022handoccnet,oh2023recovering}, or incorporate kinematic and biomechanical priors to suppress implausible poses~\cite{spurr2020weakly,xie2024ms}.

More recently, the dominant trend has been to scale model capacity and training data with transformer backbones, following their success in body pose and mesh recovery~\cite{xu2022vitpose, lin2021end}.
HandDiff~\cite{cheng2024handdiff} explores a diffusion-based formulation that generates hand poses through iterative denoising.
In contrast, HaMeR~\cite{pavlakos2024hamer} adopts a minimalist foundation-style design that fine-tunes a ViT backbone~\cite{xu2022vitpose} to directly regress MANO pose, shape, and camera from a single RGB image, trained on a mixed $\sim$2.7M-image corpus.
Hamba~\cite{dong2024hamba} replaces attention with a graph-guided Mamba~\cite{gu2024mamba} backbone to capture joint spatial relations.
WiLoR~\cite{potamias2024wilor} further scales training to $\sim$4.2M RGB images with a coarse-to-fine refinement module, achieving state-of-the-art RGB performance.
These works show that simple architectures can achieve strong performance when trained with large and diverse data.

However, RGB-based pipelines remain ill-posed in depth, leading to global translation and scale errors even when 2D projections seem well aligned.
To exploit geometric cues when depth is available, prior work has explored RGB-D formulations that regress from depth maps or multimodal inputs such as image-point-cloud hybrids~\cite{liu2023sa,ren2023ipnet}.
As a recent representative, Keypoint-Fusion~\cite{liu2024keypoint} fuses RGB and depth features around hand keypoints to reduce RGB ambiguity.
Nevertheless, the lack of large-scale, aligned RGB-D datasets has limited these methods to benchmark-specific training, hindering cross-dataset generalization.
To address this, we include aligned depth in our synthetic data, enabling transformer-based models to use geometric cues while retaining data-driven generalization.

\input{tables/synthetic_dataset_summary.tex}

\parhead{Synthetic data for 3D hand pose estimation and transfer.}
Collecting large-scale real 3D hand annotations is difficult, especially for hand-object interactions where hands are small, articulated, and frequently occluded.
This challenge is even greater for RGB-D learning, since large datasets with well-aligned depth and reliable 3D labels remain scarce.
Synthetic data therefore provides a practical way to scale supervision while obtaining paired RGB-D observations with full 3D ground truth.

As summarized in Tab.~\ref{tab:synthetic_dataset_summary}, prior synthetic hand datasets cover different aspects of this problem, including aligned depth~\cite{zimmermann2017rhd}, hand-object interaction~\cite{hasson2019obman}, and more realistic rendering with HDR backgrounds and dynamic lighting~\cite{moon2023re-interhand,li2023renderih}.
However, existing datasets rarely provide scale, realistic rendering, arm context, object-induced occlusion, and aligned depth together.
Recent analysis of the synthetic-to-real gap further suggests that forearm context, interaction realism, occlusion, and rendering variation are important factors for transfer~\cite{zhao2025analyzing}.
We provide a more detailed comparison with prior synthetic datasets in Appx.~\ref{app:synthetic_dataset_discussion}.

Overall, existing datasets rarely combine large scale, hand-object occlusion, arm context, realistic rendering variation, and RGB-aligned depth cues with rich 3D annotations.
This gap motivates a synthetic data source that can complement real RGB training while supporting depth-aware extensions.

%% file: tables/synthetic_dataset_summary.tex
\begin{table*}[b!]
\centering
\small
\vspace{-0.1in}
\caption{
    \textbf{Comparisons with representative synthetic hand-pose datasets.}
    {\datasetname} provides the most comprehensive setting among prior work, combining the largest scale with realistic HDR indoor/outdoor backgrounds, diverse dynamic lighting, aligned depth, and arm/object configurations (\datasetname-\texttt{Single}/\texttt{Interact}).
    }
\resizebox{\linewidth}{!}{
\scriptsize
\begin{tabular}{l c c c c c c} 
\toprule
\textbf{Dataset}        & \textbf{Size} & \textbf{Background}             & \textbf{Lighting} & \textbf{Depth}                       & \textbf{Arm}                         & \textbf{Object}                      \\
\midrule
RHD \cite{zimmermann2017rhd}           & 44K  & static                 & manual   & \Checkmark   & \XSolidBrush & \XSolidBrush \\
ObMan \cite{hasson2019obman}         & 154K & static                 & manual   & \Checkmark & \Checkmark   & \Checkmark   \\
Re:InterHand \cite{moon2023re-interhand} & 1.5M & HDR - Indoor           & dynamic  & \XSolidBrush & \XSolidBrush & \XSolidBrush \\
RenderIH \cite{li2023renderih}      & 1.2M   & HDR - Indoor / Ourdoor & dynamic  & \XSolidBrush & \XSolidBrush & \XSolidBrush \\
\midrule
{\datasetnamedpx} (Ours)     & 2.5M & HDR - Indoor / Outdoor & dynamic  & \Checkmark   & \Checkmark   & \XSolidBrush \\
{\datasetnamegxl} (Ours) & 4.1M & HDR - Indoor / Outdoor & dynamic  & \Checkmark   & \Checkmark   & \Checkmark   \\
\bottomrule
\end{tabular}
}
\label{tab:synthetic_dataset_summary}
\end{table*}

%% file: sec/3_data.tex
\vspace{-0.1in}
\section{\datasetname dataset}
\label{sec:synth_data}
\vspace{-0.1in}

Motivated by the need for scalable RGB-D supervision, 
we propose {\datasetname}, a large-scale synthetic dataset designed to complement existing real-world data and support both RGB-only and RGB-D model training, guided by two principles.
First, the data should be \emph{diverse} in pose, shape, appearance, viewpoint, and interaction patterns to support large-capacity models.
Second, the data should be \emph{geometrically grounded}: 
we provide depth cues and precise hand annotations obtained from the rendering and simulation pipeline.
Concretely, {\datasetname} comprises two complementary branches: {\datasetnamedpx} with $2.5$M RGB-D samples for diverse single-hand settings, and {\datasetnamegxl} with $4.1$M RGB-D samples for hand-object interactions with object-induced occlusions.
Detailed dataset statistics are provided in Appx.~\ref{sec:suppl_dataset}.



\vspace{-0.1in}
\subsection{Dataset creation}
\vspace{-0.05in}
\label{sec:dataset_creation}

\input{figs/dataset_example}

\parhead{Hand shapes.}
To cover a broad range of hand shapes, but also to avoid unrealistic geometry, we sample $47,438$ MANO shape parameters $\beta$ from the empirical distribution of FreiHAND \cite{zimmermann2019freihand} and InterHand2.6M \cite{moon2020interhand26m},
which are real datasets with large subject-level diversity and provide a good proxy for the true distribution of hand shapes in the population.

\parhead{Hand poses.}
Hand realism requires plausible articulation.
Rather than sampling poses from simple heuristics, we leverage DPoser-Hand~\cite{lu2025dposerx}, a hand pose model trained on a mixture of large real datasets (\eg FreiHAND~\cite{zimmermann2019freihand}, HO-3D~\cite{hampali2020ho3d}, DexYCB~\cite{chao2021dexycb}, H2O~\cite{kwon2021h2o}, and Re:InterHand~\cite{moon2023re-interhand}).
The \emph{key} advantage of such a diffusion prior is that it captures multiple pose distributions observed in real data, enabling us to generate a broad range of poses while avoiding unnatural articulations that often arise from naive sampling. 
During dataset generation, hand poses are sampled on the fly from this prior. 

\parhead{Hand textures.}
A key limitation of prior synthetic datasets is the lack of high-fidelity hand textures, which are crucial for closing the sim-to-real gap.
To address this,
we leverage a hand texture generator, Handy~\cite{potamias2023handy}, to produce a large variety of realistic high-frequency skin patterns,
which improves the visual fidelity of rendered creases and shading transitions compared to the canonical MANO texture space.
In particular,
we adopt Handy as our primary source of hand textures and then augment it by applying controlled color transformations, such as hue and saturation perturbations, that broaden skin-tone distribution while preserving the high-frequency texture structure.
This yields the diverse textures of $10,240$ unique hand appearances,
which is significantly more than the limited hand-crafted texture libraries used in prior works~\cite{zimmermann2017rhd, hasson2019obman, moon2023re-interhand, li2023renderih}.

\parhead{Forearm textures.}
To maintain realistic appearance continuity across the hand--forearm junction, we attach a full arm mesh from SMPL-H~~\cite{SMPL:2015, SMPL-X:2019, romero2017mano} during dataset generation. 
We texture the arm using SMPLitex~\cite{casas2023smplitex}, which provides $254$ high-quality human-body textures for realistic rendering. 
This introduces diverse and natural skin appearances for the visible forearm regions.

\parhead{Backgrounds.}
To diversify environments, we use two sources of high-resolution backgrounds: 
the MIT Indoor Scenes dataset~\cite{MIT_Indoor_2009} ($536$ images) and a pool of $734$ HDRI environment maps from Poly Haven, formerly HDRI Haven~\cite{polyhaven_hdri}.
For each rendered hand sample, a background is randomly selected, rotated, zoomed in, and then cropped with a random size and location to a patch before being used as the sample-specific background. 
This ensures the effective background appearance vary across samples, even when the same source image or HDRI map is selected multiple times.

\parhead{Lighting.}
For foreground-background consistency,
we randomize a small set of scene lights and correlate their color statistics with the background patch,
so that the rendered hand inherits the dominant illumination tone of the scene (\eg warm indoor tungsten \vs cooler daylight).
For HDRI environment maps, we directly use the environment illumination to obtain coherent scene lighting.

\parhead{Cameras.}
To ensure viewpoint diversity, we randomize camera intrinsics and extrinsics within realistic ranges based on calibration statistics from real datasets such as HO-3D~\cite{hampali2020ho3d} and DexYCB~\cite{chao2021dexycb}. 
The intrinsics, including focal length and field of view, are sampled to mimic those of the real capture setups. 
For the extrinsics, we keep the camera-to-hand distance within realistic bounds (within $0.6$-$0.8$m) and randomize the camera viewpoint over the full 3D viewing sphere around the hand, enabling observations from all-possible directions while constraining the hand to remain well framed, avoiding severe truncation artifacts.

\input{figs/fig_qualitative_concat}

\parhead{Rendering \datasetnamedpx dataset.}
After preparing the above-described components,
we instantiate them into a unified rendering-and-compositing pipeline.
We render all scenes in SAPIEN~\cite{xiang2020sapien},
using its ray-tracing renderer to better capture realistic shading, cast shadows, and specular effects.
For each sample,
we first draw a plausible MANO shape and pose,
and apply diffused hand textures with optional color perturbations to get the textured 3D hand mesh.
Then, we attach a textured forearm segment to the hand in 3D, which provides consistent geometry and appearance across the hand--forearm junction and avoids boundary artifacts common in 2D compositing.
In particular, we extract the arm mesh from a parametric body model (SMPL/SMPL+H family~\cite{SMPL:2015, SMPL-X:2019, romero2017mano}), align it to the MANO wrist frame, and texture it using SMPLitex~\cite{casas2023smplitex}.
Finally, we render the foreground hand-arm with background-aware lighting by using the above-described camera randomization. 
For each scene, we render two images from two independently sampled camera viewpoints, improving simulator efficiency while increasing perspective diversity.
To further diversify the rendered data,
we composite the rendered foreground onto a randomly cropped high-resolution background patch and pair it with HDR environment illumination,
while keeping foreground-background consistency by correlating scene-light color statistics with the sampled background patch.

\parhead{Aligned depth maps.}
We also provide aligned depth maps for all synthetic samples.
To this end,
we first render accurate metric depths for the hand and forearm from SAPIEN.
For the background patch, we estimate a dense metric depth map using MoGe-2~\cite{wang2025moge2accuratemonoculargeometry}.
We then directly fuse foreground and background depth in camera space to obtain a dense depth image for the final composite.
While this is not a ``perfect'' ground truth depth map due to differences in camera intrinsics between the rendered foreground and background, as well as noise in the estimated background depth,
it provides a useful approximation for training and evaluation purposes.
Besides,
we also store a foreground mask so models can optionally restrict losses or depth usage to valid hand/arm regions.

\parhead{Rendering \datasetnamegxl dataset.}
To model occluded hands in real-world scenarios at scale,
we render a second branch of the dataset, \texttt{AnyHand}-\texttt{Interact}, by rendering the grasping behaviors from GraspXL~\cite{zhang2024graspxl}, which provides over $10$M physics-simulation-based hand-object interaction sequences on more than $500$k realistically textured objects spanning diverse categories and surface appearances from Objaverse~\cite{deitke2023objaverse}, with contact-consistent grasps and natural occlusion patterns.
Therefore, we directly use the full GraspXL corpus and inherit its associated object set.
The rendering pipeline follows the same strategy as the single-hand branch, but now includes realistic mutual occlusions between the hand and the manipulated object.


\vspace{-0.1in}
\subsection{Dataset statistics}
\vspace{-0.1in}
Overall,
\datasetname consists of \datasetnamedpx (with $1.25$M scenes and $2.5$M images) and \datasetnamegxl (with $2.05$M scenes and $4.1$M images),
which are rendered with a combination of $47,438$ hand shapes, $10,240$ hand textures, $254$ arm textures, $1,270$ backgrounds, and more than 500k objects from Objaverse~\cite{deitke2023objaverse}.
Note that, for all rendered samples, we store RGB, depth, foreground mask, 2D bounding boxes, together with camera intrinsics and extrinsics.
We also provide precise 3D hand pose and shape parameters directly from the simulation, which can be used for supervised training or evaluation.
More details on the implementation can be found in Appx. \ref{subsec:sup_data_gen}.


%% file: figs/dataset_example.tex
\begin{figure*}[b!]
    \centering
    \vspace{-0.1in}
    \includegraphics[width=0.75\linewidth]{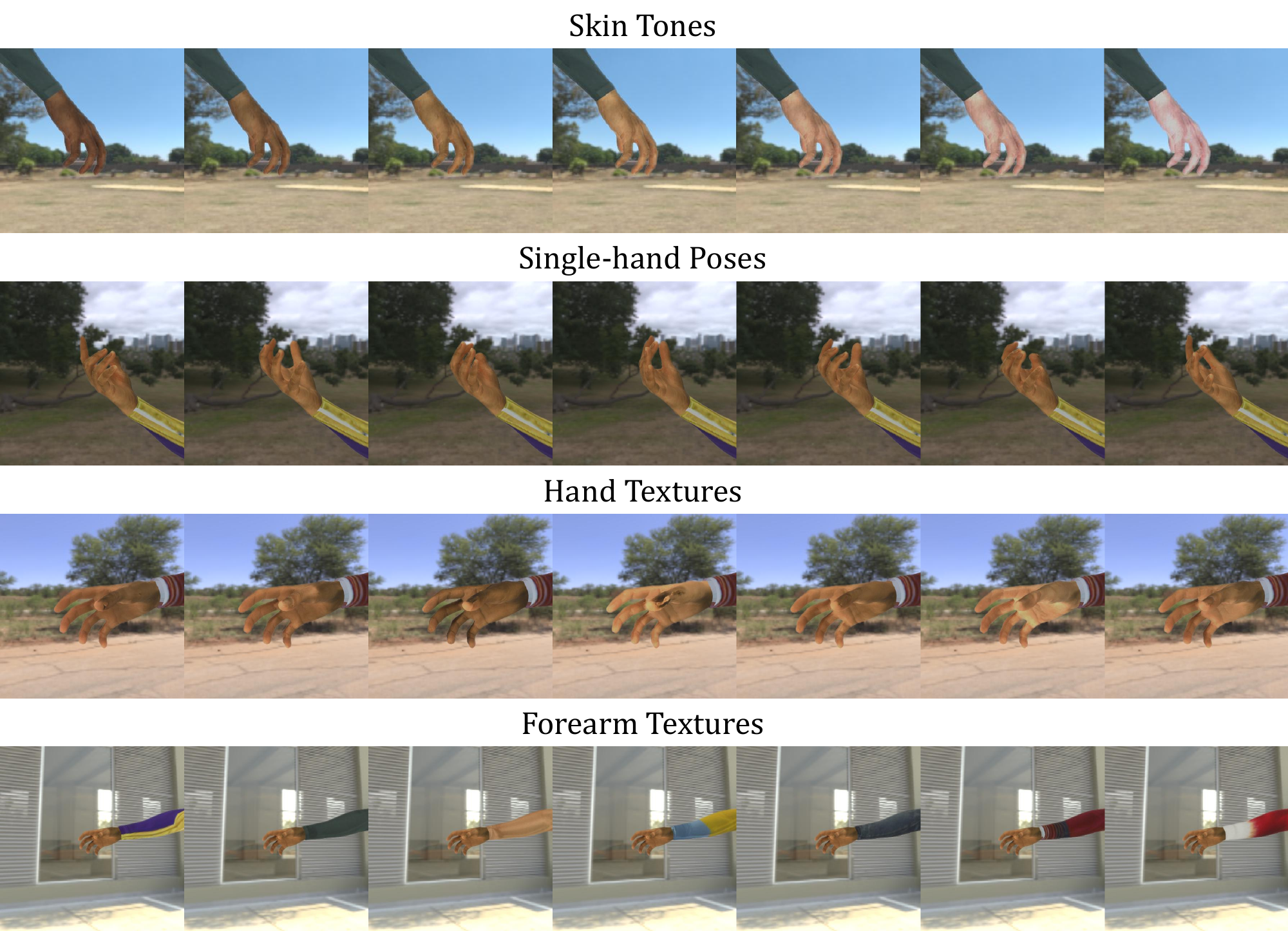}
    \caption{\textbf{Qualitative visualizations of controllable variations.} 
    We show representative samples from our generator by varying one factor at a time: 
    \emph{skin tones}, 
    \emph{single-hand poses} from DPoser-Hand~\cite{lu2025dposerx}, 
    \emph{hand textures}  from Handy~\cite{potamias2023handy}, 
    and \emph{forearm appearance} from SMPLitex~\cite{casas2023smplitex}. 
    These examples demonstrate the diversity of appearance and context that we leverage in {\datasetname}. 
    }
    \label{fig:dpx-ablation-vis}
\end{figure*}

%% file: figs/fig_qualitative_concat.tex
\begin{figure*}[b!]
    \centering
    \vspace{-0.2in}    \includegraphics[width=1.0\linewidth]{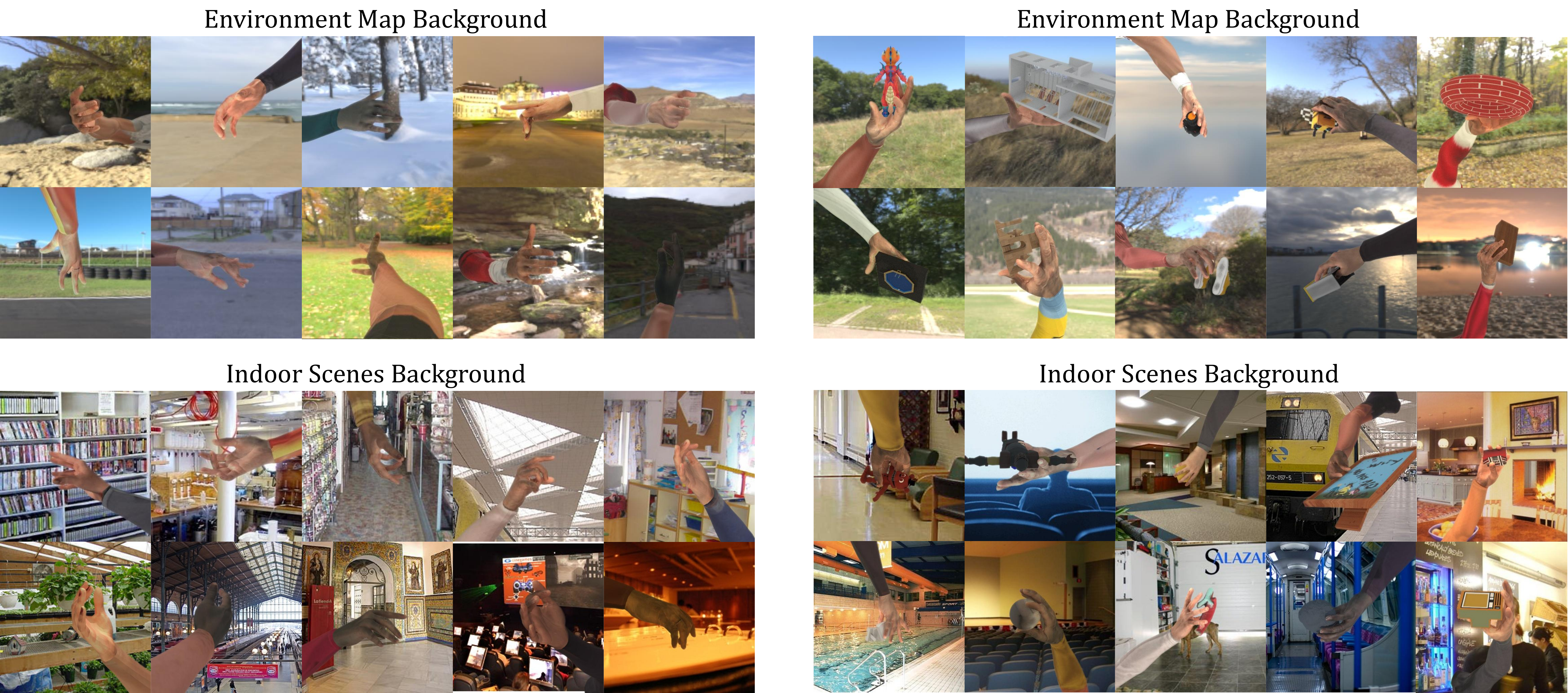}
    \caption{\textbf{Qualitative visualizations.} Examples of {\datasetnamedpx} (left) and {\datasetnamegxl} (right), with both HDR environment-map backgrounds (top) and real indoor scenes (bottom). 
    In addition to diverse hand/arm appearance and poses, we have additional diversity on the interacted objects and grasp configurations, producing a wide range of object-induced hand occlusions and self-occlusions under varying perspectives. }
    \label{fig:qualitative_vis_concat}
\end{figure*}

%% file: sec/4_RGB_exp.tex
\vspace{-0.1in}
\section{Assessing {\datasetname} dataset on RGB-only settings}
\label{sec:experiments}
\vspace{-0.1in}

\subsection{Experiment}
\vspace{-0.05in}

\parhead{Evaluation setups.}
To evaluate the quality of {\datasetname} and its effectiveness for improving foundation-style hand mesh reconstruction, we study two representative frameworks: 
HaMeR~\cite{pavlakos2024hamer}, trained on 2.7M images, and WiLoR~\cite{potamias2024wilor}, trained on 4.2M images. 
For both frameworks, we augment their original training corpora with {\datasetname} while keeping the model architectures and training hyper-parameters unchanged. 
To isolate dataset design from simply adding synthetic data, we also train HaMeR with RenderIH~\cite{li2023renderih}, a representative large-scale synthetic hand dataset containing 1.2M images with HDR backgrounds and dynamic lighting, using the same mixed-training protocol.

Each training batch is constructed by sampling from different datasets according to predefined weights. 
We assign a weight of $0.4$ to the added synthetic dataset and scale the weights of the original datasets by $0.6$, preserving the original data distribution while introducing substantial synthetic supervision. 
Details on the mixed-training scheme are provided in Appx.~\ref{subsec:training_corpora_details}. 
All experiments run on $8$ NVIDIA H100 GPUs and are trained for 5--7 days until convergence. 
Compute details are provided in Appx.~\ref{app:detail}.

\input{tables/table_part4_freihand}
\input{tables/table_part4_ho3d}

\parhead{Metrics.}
Following protocols in the original papers of HaMeR\cite{pavlakos2024hamer} and WiLoR\cite{potamias2024wilor}, we report Procrustes-aligned Mean per Joint and Vertex Error (PA-MPJPE, PA-MPVPE) and the F-score of vertices at $5$mm and $15$mm (F@5, F@15)~\cite{zimmermann2019freihand, knapitsch2017tanks}.
For the HO-3D\cite{hampali2020ho3d} dataset, we additionally report AUC$_j$ and AUC$_v$, defined as area under the PCK curve over joint and vertex error thresholds.

\vspace{-0.1in}
\subsection{Results}
\vspace{-0.1in}


\parhead{In-domain results.}
\label{sec:mix_train}
We perform an in-domain evaluation on the popular FreiHAND~\cite{zimmermann2019freihand} and HO-3D v2~\cite{hampali2020ho3d} benchmarks.
As shown in Tab.~\ref{tab:real_syn_freihand} and \ref{tab:real_syn_ho3d_v2},
adding {\datasetname} consistently improves performance across all metrics for both HaMeR and WiLoR, demonstrating the effectiveness of our synthetic data for enhancing RGB-based hand pose estimation.
On FreiHAND, WiLoR w/ {\datasetname} achieves the best results,
while the effect of synthetic augmentation is even more pronounced for HaMeR.
The PA-MPJPE drops from $6.0$\,mm to $5.54$\,mm 
, and PA-MPVPE decreases from $5.7$\,mm to $5.24$\,mm 
, lifting HaMeR into the same performance tier as the top-ranked approaches,
\emph{without requiring any specific architecture modification}.
On HO-3D v2, which emphasizes hand-object interactions, the same trend holds. 
WiLoR w/ {\datasetname} attains the best overall results, while HaMeR also improves substantially, with PA-MPJPE of HaMeR w/ {\datasetname} reduced 
from $7.7$\,mm to $7.47$\,mm. 
In contrast, training with RenderIH~\cite{li2023renderih} under the same protocol yields smaller improvements on both benchmarks, highlighting the quality and effectiveness of {\datasetname}. 
Qualitative examples in Appx.~Fig.~\ref{fig:qualitative_vis_freihand} further illustrate that training with {\datasetname} helps reduce errors in challenging poses and occluded joints, which are common failure modes for RGB-based methods.
More visual results are in Appx.~\ref{subsec:sup_qualitative_results}.


\input{tables/table_part4_hocap}

\parhead{Out-of-domain results.}
To evaluate the out-of-domain generalization ability,
we directly evaluate performance on the HO-Cap~\cite{wang2024hocap} benchmark without any fine-tuning,
whose images are entirely from unseen sources with a clear domain shift.
As reported in Tab.~\ref{tab:real_syn_hocap},
augmenting training with {\datasetname}'s synthetic data improves both WiLoR and HaMeR,
and yields a notable ranking change: HaMeR w/ {\datasetname} attains PA-MPJPE of 4.66\,mm, slightly outperforming WiLoR w/ {\datasetname}, which has PA-MPJPE of 4.69\,mm.
This is particularly interesting because HaMeR is overall weaker than WiLoR on HO-Cap under the original training setup.
The fact suggests that the performance gap between the two architectures is not fixed, but can be influenced by the training data distribution.

\parhead{Analysis.}
Two key observations emerge from these experiments.
Firstly, adding \datasetname's synthetic data in the original training data corpora provides consistent improvements across two representative ViT-based baselines (HaMeR and WiLoR), suggesting that the gains are not tied to a single pipeline design.
The stronger gains on HaMeR are consistent with its smaller training dataset, whereas WiLoR’s additional refinements and larger training corpus leave less room for improvement.
Moreover, on HO-Cap, the performance gains from adding \datasetname\ are substantially larger than the differences between architectures, suggesting that improvements from training data can outweigh architectural differences.
Overall, these findings support our thesis that scaling training data quality, quantity, and diversity is a stronger lever than iterating on architectures alone.

\input{figs/fig_hamer_scaling}

\vspace{-0.1in}
\subsection{Ablations on {\datasetname}}
\label{subsec:main_ablations}
\vspace{-0.1in}

We ablate our \datasetname for in-domain and out-of-domain performance in Fig.~\ref{fig:hamer_scaling_curves} and Tab.~\ref{tab:part4_ablations}.
We keep the HaMeR~\cite{pavlakos2024hamer} architecture and training protocol fixed, and only vary the data configurations.

\parhead{Scaling of adding {\datasetname} in training.}
Fig.~\ref{fig:hamer_scaling_curves} studies the effect of synthetic-data scaling by varying the amount of synthetic data while keeping the real-data portion fixed.
The results show that augmenting HaMeR with {\datasetname} yields a substantial performance boost over the no-synthetic baseline across all benchmarks.
On the in-domain benchmarks (FreiHAND and HO-3D v2),
the small dataset ($\frac{1}{3}$ on FreiHAND, $\frac{2}{3}$ on HO-3D) already provides the main gains,
while increasing it to full size only provides a modest improvement.
However, on the out-of-domain HO-Cap benchmark, we see a more consistent improvement as the synthetic budget increases, suggesting that scaling up synthetic data may be particularly beneficial for improving robustness to domain shifts.
This is likely because the larger synthetic dataset covers a wider range of poses, shapes, textures, and backgrounds that better match the diversity of \emph{unseen} real-world scenarios.
This provides a strong motivation for investing in large-scale synthetic data generation in future work.

\parhead{Sim-and-real mixed-training recipe.}
The scaling experiment above shows that adding \datasetname improves performance, but it does not answer how synthetic and real data should be mixed. 
To study this, we fix the total training set size to 2.7M samples, matching HaMeR's original training budget, and vary the proportion of \datasetname. 
For example, when \datasetname accounts for 25\% of the 2.7M data, the remaining 75\% samples are drawn from HaMeR's original training corpus and scaled down proportionally. 
All models are trained for the same number of steps for a controlled comparison.

As shown in Fig.~\ref{fig:hamer_div_scaling}, using \datasetname alone is not sufficient: the 100\% \datasetname setting performs worse than the mixed settings and also degrades relative to HaMeR's original recipe. 
This confirms that \datasetname is best used as complementary supervision rather than as a full replacement for real training data. 
In contrast, all mixed settings improve over the 0\% baseline, showing that \datasetname consistently benefits HaMeR under the same training budget. 
Notably, the 50\% and 75\% mixtures achieve strong performance across the evaluated benchmarks, and in several cases surpass the WiLoR reference despite using the simpler HaMeR architecture and less training data, suggesting that mixing a large proportion of synthetic data with real data is an effective strategy for improving hand pose estimation.


\label{subsec:ablation_cotrain_recipe}
\input{figs/fig_hamer_div_scaling}


\input{tables/table_part4_ablations}

\parhead{Ablations on other variants.}
As summarized in Tab.~\ref{tab:part4_ablations}, we further ablate key design choices in {\datasetname}.
First, dropping either the \texttt{Single} branch or the \texttt{Interact} branch degrades performance, suggesting that when the target benchmarks contain a mix of single-hand and hand-object interaction cases, jointly training with both single and interaction samples yields the best overall results.
Second, removing arm texture slightly hurts performance, suggesting that realistic arm appearance (beyond geometry alone) provides useful contextual cues and improves generalization.
Third, replacing diffusion-based pose synthesis with poses interpolated from real data leads to a consistent performance drop, indicating that diffusion provides more effective pose diversity for this mixed training scheme.

\parhead{Summary.} These ablations collectively show that the gains from \datasetname are not simply due to increased training data. Scaling synthetic data shows diminishing returns in-domain, while still improving out-of-domain performance. The mixing experiment indicates that synthetic data complements rather than replaces real data. In addition, the component ablations demonstrate that specific design choices contribute to performance. 
Overall, these findings suggest 
that both data quality and quantity contribute to the improvements in Sec. \ref{sec:mix_train}. 

%% file: tables/table_part4_freihand.tex
\begin{table*}[t!]
\centering
\caption{
   \textbf{Comparison with the state-of-the-art on the FreiHAND benchmark \cite{zimmermann2019freihand}.} 
   Top results are emphasized in \colorbox{red!30}{top1}, \colorbox{orange!30}{top2}, and \colorbox{yellow!30}{top3}.
   Notably, augmented training with \datasetname yields a \textbf{7.6\%} PA-MPJPE improvement for HaMeR and a \textbf{1.9\%} improvement for WiLoR.
   A full comparison with more prior works can be found in Appx. in Tab.~\ref{tab:suppl_rgb_freihand}.
}
\small
\begin{tabular}{@{}lcccc@{}}
\toprule
\textbf{Method}  & \textbf{PA-MPJPE} $\downarrow$ & \textbf{PA-MPVPE} $\downarrow$ & \textbf{F@5mm} $\uparrow$ & \textbf{F@15mm} $\uparrow$ \\
\midrule
AMVUR \cite{jiang2023probabilistic}$_\text{\scriptsize CVPR23}$          & 6.2      & 6.1      & 0.767                 & 0.987                  \\
HaMeR \cite{pavlakos2024hamer}$_\text{\scriptsize CVPR24}$          & 6.0      & 5.7      & 0.785                 & 0.990                  \\
Hamba \cite{dong2024hamba}$_\text{\scriptsize NeurIPS24}$         & 5.7      & 5.3      & 0.806                 & \colorbox{yellow!30}{0.992}                  \\
WiLoR \cite{potamias2024wilor}$_\text{\scriptsize CVPR25}$         & \colorbox{orange!30}{5.5}      & \colorbox{orange!30}{5.1}      & \colorbox{orange!30}{0.825}                 & \colorbox{orange!30}{0.993}                  \\
\midrule
HaMeR w/ RenderIH & $5.8_{57}$ & $5.4_{86}$ & $0.797$ & \colorbox{yellow!30}{0.992} \\
HaMeR w/ {\datasetname} (Ours)           & \colorbox{yellow!30}{$5.5_{45}$}  & \colorbox{yellow!30}{$5.2_{46}$}  & \colorbox{yellow!30}{0.811}                 & \colorbox{orange!30}{0.993}         \\
WiLoR w/ {\datasetname}  (Ours)            & \colorbox{red!30}{$5.3_{94}$}  & \colorbox{red!30}{$5.0_{46}$}  & \colorbox{red!30}{0.827}                 & \colorbox{red!30}{0.994}   \\
\bottomrule
\end{tabular}
\label{tab:real_syn_freihand}
\end{table*}

%% file: tables/table_part4_ho3d.tex
\begin{table*}[t!]
\centering
\caption{
   \textbf{Comparison with the state-of-the-art on the HO-3D v2 benchmark \cite{hampali2020ho3d}.} 
   Top results are emphasized in \colorbox{red!30}{top1}, \colorbox{orange!30}{top2}, and \colorbox{yellow!30}{top3}.
   Using \datasetname reduces PA-MPJPE by \textbf{3.0\%} for HaMeR and by \textbf{1.9\%} for WiLoR. 
   A full comparison of more prior works can be found in Appx. Tab.~\ref{tab:suppl_rgb_ho3d}.
}
\small
{\begin{tabular}{@{}lcccccc@{}}
\toprule
\textbf{Method}         & \textbf{AUC\textsubscript{j}} $\uparrow$ & \textbf{PA-MPJPE} $\downarrow$ & \textbf{AUC\textsubscript{v}} $\uparrow$ & \textbf{PA-MPVPE} $\downarrow$ & \textbf{F@5} $\uparrow$ & \textbf{F@15} $\uparrow$ \\
\midrule
AMVUR \cite{jiang2023probabilistic}          & 0.835                   & 8.3                     & 0.836                   & 8.2                     & 0.608            & 0.965             \\
HaMeR \cite{pavlakos2024hamer}        & 0.846                   & 7.7                     & 0.841                   & 7.9                     & 0.635            & 0.980             \\
Hamba \cite{dong2024hamba}       & \colorbox{yellow!30}{0.850}                   & \colorbox{yellow!30}{7.5}                     & 0.843                   & \colorbox{yellow!30}{7.7}                     & \colorbox{orange!30}{0.648}            & \colorbox{yellow!30}{0.982}             \\
WiLoR \cite{potamias2024wilor}        & \colorbox{orange!30}{0.851}                   & \colorbox{yellow!30}{7.5}                     & \colorbox{yellow!30}{0.846}                   & \colorbox{yellow!30}{7.7}                      & \colorbox{yellow!30}{0.646}            & \colorbox{orange!30}{0.983}             \\
\midrule
HaMeR w/ RenderIH & $0.846$ & $7.6_{85}$ & $0.842$ & $7.9_{02}$ & $0.634$ & \colorbox{yellow!30}{0.982} \\
HaMeR w/ {\datasetname}  (Ours)          &  \colorbox{orange!30}{0.851}                   & \colorbox{orange!30}{$7.4_{71}$}             & \colorbox{orange!30}{0.847}                   & \colorbox{orange!30}{$7.6_{76}$}           & 0.645            & \colorbox{red!30}{0.984}             \\
WiLoR w/ {\datasetname}  (Ours)          & \colorbox{red!30}{0.853}                   & \colorbox{red!30}{$7.3_{55}$}             & \colorbox{red!30}{0.848}                   & \colorbox{red!30}{$7.6_{24}$}           & \colorbox{red!30}{0.649}            & \colorbox{red!30}{0.984}             \\
\bottomrule
\end{tabular}}

\label{tab:real_syn_ho3d_v2}
\end{table*}

%% file: tables/table_part4_hocap.tex
\begin{table*}[t]
\centering
\caption{
   \textbf{Comparison with HaMeR \cite{pavlakos2024hamer} and WiLoR \cite{potamias2024wilor} on the HO-Cap benchmark \cite{wang2024hocap} as an in-the-wild case.} 
    Better results are \textbf{bolded}.
}
\input{tables/real_syn_hocap}
\vspace{-0.2in}
\label{tab:real_syn_hocap}
\end{table*}

%% file: tables/real_syn_hocap.tex
\centering
\small
\setlength{\tabcolsep}{7pt} 
\
\begin{tabular}{@{}lcccc@{}}
\toprule
\textbf{Method}         & \textbf{AUC\textsubscript{j}} $\uparrow$ & \textbf{PA-MPJPE} $\downarrow$ & \textbf{F@5mm} $\uparrow$ & \textbf{F@15mm} $\uparrow$ \\
\midrule
HaMeR \cite{pavlakos2024hamer} & 0.901 & 4.94  & 0.621 & 0.984          \\
WiLoR \cite{potamias2024wilor} & 0.899 & 5.02  & 0.615 & 0.981          \\
\midrule
HaMeR w/ {\datasetname} (Ours)          & \textbf{0.907} & \textbf{4.66}  & \textbf{0.662} & 0.985           \\
WiLoR w/ {\datasetname} (Ours)         & 0.906 & 4.69  & 0.652 & \textbf{0.987}           \\
\bottomrule
\end{tabular}

%% file: figs/fig_hamer_scaling.tex
\begin{figure*}[b!]
    \centering
    \vspace{-0.1in}
    \includegraphics[width=1.0\linewidth]{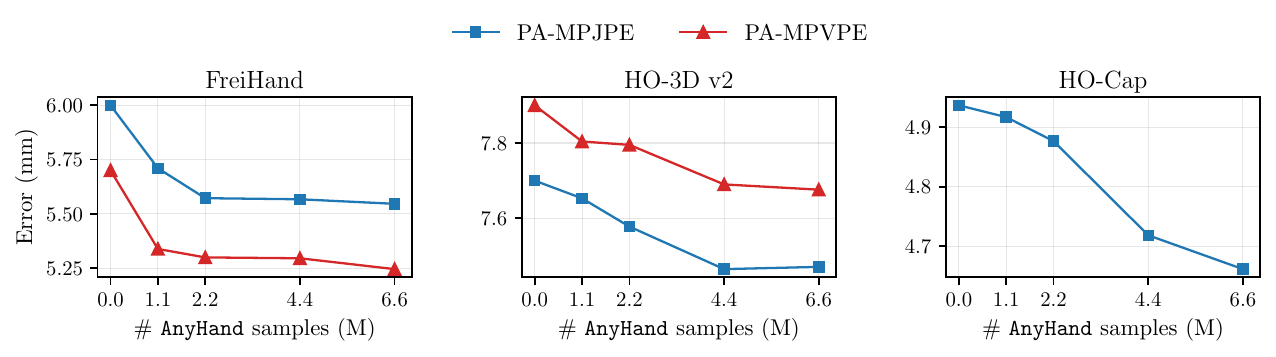}
    

    \caption{\textbf{Scaling HaMeR training with \datasetname in addition.} 
    We retrain HaMeR \cite{pavlakos2024hamer} while keeping its original real-data training set fixed, and vary the number of additional \datasetname samples used. 
    We report PA-MPJPE and PA-MPVPE on FreiHAND~\cite{zimmermann2019freihand}, HO-3D v2~\cite{hampali2020ho3d}, and HO-Cap~\cite{wang2024hocap}. 
    Adding {\datasetname} in training consistently reduces error, with gains saturating beyond $\sim$2--4M samples.}
    \label{fig:hamer_scaling_curves}
\end{figure*}

%% file: figs/fig_hamer_div_scaling.tex
\begin{figure*}[tb!]
    \centering

    \includegraphics[width=1.0\linewidth]{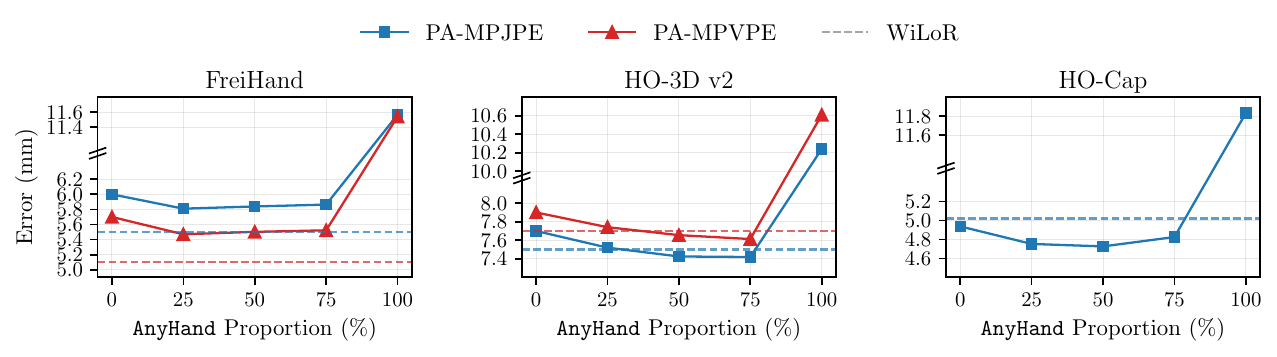}

    \caption{\textbf{Effect of the training-data mixing ratio.}
    HaMeR is trained with a fixed budget of 2.7M samples, matching its original data recipe, while varying the proportion of \datasetname. 
    The remaining samples are drawn from HaMeR's original training corpus and scaled down proportionally. 
    The dashed line indicates WiLoR's original result as a comparison. 
    The results provide a rough estimate of the impact of different mixture ratios under a fixed training budget.}
    \label{fig:hamer_div_scaling}
\end{figure*}

%% file: tables/table_part4_ablations.tex
\begin{table}[t]
\centering
\small
\caption{\textbf{Ablations on {\datasetname} components on the FreiHAND~\cite{zimmermann2019freihand} benchmark.}
In each experiment, the HaMeR~\cite{pavlakos2024hamer} model is trained with different data configurations while other settings are identical. Here ``w/ interpolated pose'' means using interpolated poses from real datasets instead of diffusing from the default DPoser-Hand~\cite{lu2025dposerx}.
}
\small
\begin{tabular}{@{}lcccc@{}}
\toprule
\textbf{HaMeR~\cite{pavlakos2024hamer} Variant}           & \textbf{PA-MPJPE} $\downarrow$ & \textbf{PA-MPVPE} $\downarrow$ & \textbf{F@5mm} $\uparrow$ & \textbf{F@15mm} $\uparrow$ \\
\midrule
HaMeR \cite{pavlakos2024hamer} & 6.000 & 5.700 & 0.7850 & 0.9900 \\
\midrule
w/ {\datasetname} & 5.545 & 5.246 & 0.8110 & 0.9930 \\
\midrule
w/ {\datasetnamegxl} & 5.850 & 5.525 & 0.7940 & 0.9919 \\
\midrule
w/ {\datasetnamedpx} & 5.626 & 5.286 & 0.8109 & 0.9930 \\
\qquad   w/o realistic arm textures & 5.686 & 5.312 & 0.8096 & 0.9926 \\
\qquad   w/ interpolated poses & 5.707 & 5.335 & 0.8056 & 0.9926 \\
\bottomrule
\end{tabular}
\label{tab:part4_ablations}
\end{table}

%% file: sec/6_conclusion.tex
\vspace{-0.15in}
\section{Conclusions}%
\label{sec:conclusions}
\vspace{-0.1in}

We have introduced {\datasetname},
a large-scale synthetic dataset that provides diverse hand scenes with rich annotations.
By incorporating \datasetname into the original training sets of state-of-the-art hand pose estimation models, we demonstrated consistent improvements across multiple benchmarks, validating the effectiveness of our synthetic data for enhancing RGB-based approaches.
We have also proposed a novel RGB-D architecture that incorporates a lightweight depth fusion module, and showed that it outperforms prior RGB-D approaches on hand pose estimation.
Overall, our results open a promising avenue for advancing hand pose estimation by improving the \emph{quality, diversity, and modality coverage} of training data, rather than solely focusing on architectural innovations.

%% file: sup_sec/rgb_exp.tex
\section{More evaluations of \datasetname on RGB-only settings}
\label{sec:suppl_rgb_exp}

\subsection{Training corpora and synthetic data proportions}
\label{subsec:training_corpora_details}

We provide additional details on the original training corpora used by the two evaluated frameworks.
HaMeR~\cite{pavlakos2024hamer} is trained on approximately $2.72$M samples, which are predominantly real images, with only about $61$K synthetic samples from RHD~\cite{zimmermann2017rhd}, corresponding to roughly $2.3\%$ of its training data.
WiLoR~\cite{potamias2024wilor} expands the training corpus to approximately $4.19$M samples by including additional datasets such as ARCTIC~\cite{fan2023arctic}, BEDLAM~\cite{black2023bedlam}, ReInterHand~\cite{moon2023re-interhand}, and HOT3D~\cite{banerjee2024hot3d}.
However, synthetic data still accounts for a relatively small portion of its training corpus, around $18.7\%$.

After enhancing the original training corpora with {\datasetname}, the proportion of synthetic data increases substantially.
For HaMeR~\cite{pavlakos2024hamer}, adding {\datasetname} raises the synthetic-data proportion from approximately $2.3\%$ to $71.5\%$.
For WiLoR~\cite{potamias2024wilor}, the proportion increases from approximately $18.7\%$ to $68.4\%$.
This represents, to the best of our knowledge, the first attempt to train transformer-based foundation-style hand mesh reconstruction models with such a large proportion of synthetic data.
These results therefore provide an important empirical study of whether large-scale synthetic RGB-D supervision can effectively improve modern hand pose estimators.

\subsection{Qualitative comparisons}
\label{subsec:sup_qualitative_results}
\input{figs/fig_freihand_vis}
\input{figs/fig_suppl_in_the_wild}

Fig.~\ref{fig:qualitative_vis_freihand} provides qualitative comparisons between the original WiLoR and WiLoR trained with \datasetname on FreiHAND and \datasetname test examples. 
The improvements are most visible in challenging cases involving bent fingers, unusual hand poses, and partial occlusions. 
Compared with the original WiLoR, WiLoR w/ \datasetname better matches the image evidence around fingertips and finger joints, producing meshes with more accurate finger bending and hand shape. 
These examples support our quantitative results and show that \datasetname helps reduce common RGB-based failure cases, such as over-smoothed fingers, inaccurate joint angles, and poor alignment under occlusion.

To further illustrate the qualitative prediction quality of models trained with \datasetname in addition, we provide additional visual comparisons in Fig.~\ref{fig:suppl_in_the_wild}. 
The figure includes seven real-image examples from the HO-Cap~\cite{wang2024hocap} dataset, two from the HO-3D evaluation set~\cite{hampali2020ho3d}, and one in-the-wild web image. 
These examples allow us to examine model behavior on both standard benchmark data and less controlled real-world imagery.

As shown in Fig.~\ref{fig:suppl_in_the_wild}, WiLoR w/ \datasetname more consistently recovers plausible palm width, finger thickness, and overall hand extent compared with the original WiLoR and HaMeR. 
The improvement is especially clear in hand-object interaction cases, where WiLoR w/ \datasetname better preserves fine articulation details such as finger bending angles, fingertip placement, and the projected contour of the hand. 
Overall, these visualizations suggest that training with \datasetname not only improves standard pose-estimation accuracy, but also helps produce more realistic meshes with better image-space alignment in challenging real-world cases.

\subsection{Full performance on FreiHAND}

\input{tables/table_suppl_freihand_full}

We report the complete version of Tab.~\ref{tab:real_syn_freihand} from the main paper in Tab.~\ref{tab:suppl_rgb_freihand}.

\subsection{Full performance on HO-3D}

We report the complete version of Tab.~\ref{tab:real_syn_ho3d_v2} from the main paper in Tab.~\ref{tab:suppl_rgb_ho3d}.

\input{tables/table_suppl_ho3d_full}


%% file: figs/fig_freihand_vis.tex
\begin{figure*}[bth!]
    \centering
    \includegraphics[width=\linewidth]{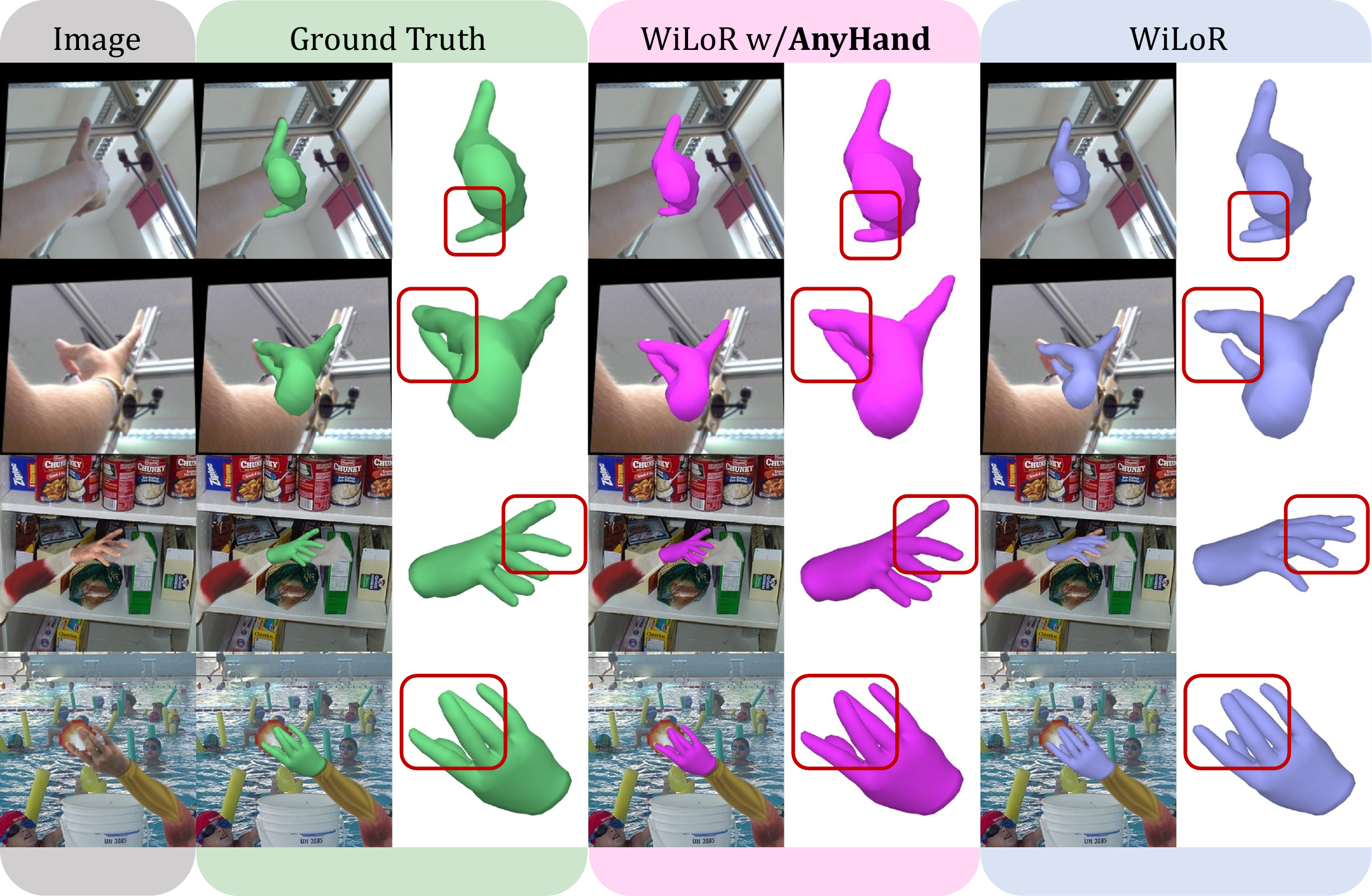}
    \caption{
    \textbf{Qualitative visualizations.} WiLoR w/ {\datasetname} vs.\ WiLoR on FreiHAND~\cite{zimmermann2019freihand} and \datasetname Test Set. 
    Left to right: input, GT, WiLoR w/ {\datasetname}, WiLoR.
    Adding synthetic data improves fine-grained pose estimation, particularly fingertip bending and finger joint angles (as boxed), yielding meshes that better match the image evidence. 
    }
    \label{fig:qualitative_vis_freihand}
\end{figure*}

%% file: figs/fig_suppl_in_the_wild.tex
\begin{figure*}[t!]
    \centering
    \includegraphics[width=0.88\linewidth]{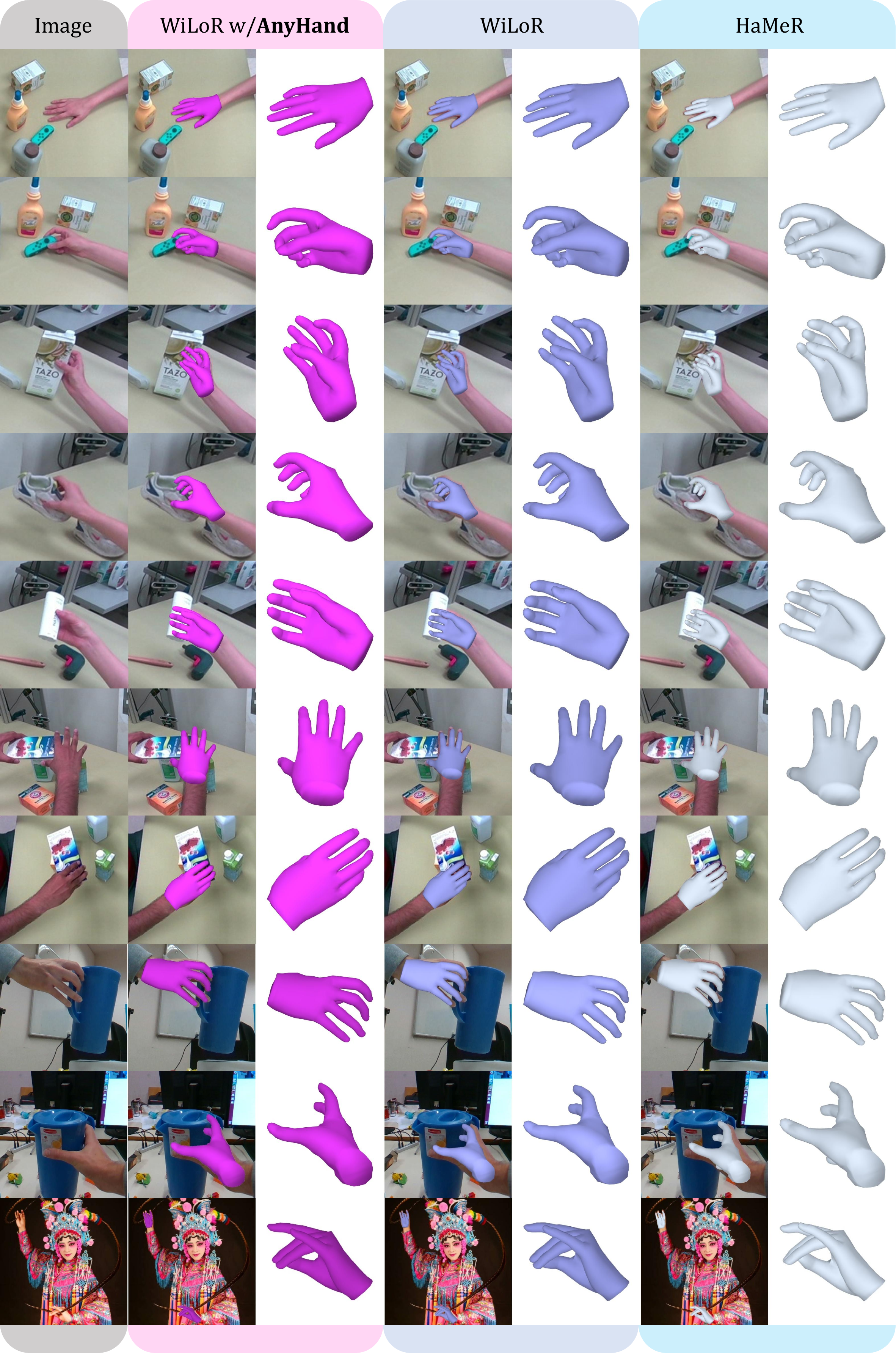}
    \caption{\textbf{Additional in-the-wild qualitative comparisons.} We compare WiLoR trained with \datasetname against the original WiLoR and HaMeR on more real-world images. WiLoR w/\datasetname shows better mesh-to-image alignment, more accurate hand scale and shape recovery, and more faithful articulation for challenging hand-object interactions and viewpoint changes.}
    \label{fig:suppl_in_the_wild}
\end{figure*}

%% file: tables/table_suppl_freihand_full.tex
\begin{table*}[t!]
\centering
\caption{
   \textbf{Comparison with the state-of-the-art on the FreiHAND benchmark \cite{zimmermann2019freihand}.} 
   Top results are emphasized in \colorbox{red!30}{top1}, \colorbox{orange!30}{top2}, and \colorbox{yellow!30}{top3}.
   Notably, augmented training with \datasetname yields a \textbf{7.6\%} PA-MPJPE improvement for HaMeR and a \textbf{1.9\%} improvement for WiLoR.
}
\small
\begin{tabular}{@{}lcccc@{}}
\toprule
\textbf{Method}  & \textbf{PA-MPJPE} $\downarrow$ & \textbf{PA-MPVPE} $\downarrow$ & \textbf{F@5mm} $\uparrow$ & \textbf{F@15mm} $\uparrow$ \\
\midrule
I2L-MeshNet \cite{moon2020i2l}     & 7.4      & 7.6      & 0.681                 & 0.973                  \\
Pose2Mesh \cite{choi2020pose2mesh}      & 7.7      & 7.8      & 0.674                 & 0.969                  \\
I2UV-HandNet \cite{chen2021i2uv}   & 6.7      & 6.9      & 0.707                 & 0.977                  \\
METRO \cite{lin2021end}          & 6.5      & 6.3      & 0.731                 & 0.984                  \\
Mesh Graphormer \cite{lin2021mesh} & 5.9      & 6.0      & 0.764                 & 0.986                  \\
MobRecon \cite{chen2022mobrecon}       & 5.7      & 5.8      & 0.784                 & 0.986                  \\
AMVUR \cite{jiang2023probabilistic}$_\text{\scriptsize CVPR23}$          & 6.2      & 6.1      & 0.767                 & 0.987                  \\
HaMeR \cite{pavlakos2024hamer}$_\text{\scriptsize CVPR24}$          & 6.0      & 5.7      & 0.785                 & 0.990                  \\
Hamba \cite{dong2024hamba}$_\text{\scriptsize NeurIPS24}$         & 5.7      & 5.3      & 0.806                 & \colorbox{yellow!30}{0.992}                  \\
WiLoR \cite{potamias2024wilor}$_\text{\scriptsize CVPR25}$         & \colorbox{orange!30}{5.5}      & \colorbox{orange!30}{5.1}      & \colorbox{orange!30}{0.825}                 & \colorbox{orange!30}{0.993}                  \\
\midrule
HaMeR w/ {\datasetname}            & \colorbox{yellow!30}{$5.5_{45}$}  & \colorbox{yellow!30}{$5.2_{46}$}  & \colorbox{yellow!30}{0.811}                 & \colorbox{orange!30}{0.993}         \\
WiLoR w/ {\datasetname}            & \colorbox{red!30}{$5.3_{94}$}  & \colorbox{red!30}{$5.0_{46}$}  & \colorbox{red!30}{0.827}                 & \colorbox{red!30}{0.994}   \\
\bottomrule
\end{tabular}
\label{tab:suppl_rgb_freihand}
\end{table*}

%% file: tables/table_suppl_ho3d_full.tex
\begin{table*}[hbt!]
\centering
\caption{
   \textbf{Comparison with the state-of-the-art on the HO-3D v2 benchmark \cite{hampali2020ho3d}.} 
   Top results are emphasized in \colorbox{red!30}{top1}, \colorbox{orange!30}{top2}, and \colorbox{yellow!30}{top3}.
   Using \datasetname reduces PA-MPJPE by \textbf{3.0\%} for HaMeR and by \textbf{1.9\%} for WiLoR. 
}
\small
{\begin{tabular}{@{}lcccccc@{}}
\toprule
\textbf{Method}         & \textbf{AUC\textsubscript{j}} $\uparrow$ & \textbf{PA-MPJPE} $\downarrow$ & \textbf{AUC\textsubscript{v}} $\uparrow$ & \textbf{PA-MPVPE} $\downarrow$ & \textbf{F@5} $\uparrow$ & \textbf{F@15} $\uparrow$ \\
\hline
Liu \etal \cite{liu2021semi}    & 0.803                   & 9.9                     & 0.810                   & 9.5                     & 0.528            & 0.956             \\
HandOccNet \cite{park2022handoccnet}   & 0.819                   & 9.1                     & 0.819                   & 8.8                     & 0.564            & 0.963             \\
I2UV-HandNet \cite{chen2021i2uv}  & 0.804                   & 9.9                     & 0.799                   & 10.1                    & 0.500            & 0.943             \\
HOnnotate \cite{hampali2020honnotate} & 0.788                   & 10.7                    & 0.790                   & 10.6                    & 0.506            & 0.942             \\
Hasson \etal \cite{hasson2019learning}  & 0.780                   & 11.0                    & 0.777                   & 11.2                    & 0.464            & 0.939             \\
ArtiBoost \cite{yang2022artiboost}    & 0.773                   & 11.4                    & 0.782                   & 10.9                    & 0.488            & 0.944             \\
Pose2Mesh \cite{choi2020pose2mesh}    & 0.754                   & 12.5                    & 0.749                   & 12.7                    & 0.441            & 0.909             \\
I2L-MeshNet  \cite{moon2020i2l}   & 0.775                   & 11.2                    & 0.722                   & 13.9                    & 0.409            & 0.932             \\
METRO  \cite{lin2021end}       & 0.792                   & 10.4                    & 0.779                   & 11.1                    & 0.484            & 0.946             \\
MobRecon \cite{chen2022mobrecon}     & -                       & 9.2                     & -                       & 9.4                     & 0.538            & 0.957             \\
KeypointTrans \cite{hampali2022keypoint} & 0.786                   & 10.8                    & -                       & -                       & -                & -                 \\
AMVUR \cite{jiang2023probabilistic}          & 0.835                   & 8.3                     & 0.836                   & 8.2                     & 0.608            & 0.965             \\
HaMeR \cite{pavlakos2024hamer}        & 0.846                   & 7.7                     & 0.841                   & 7.9                     & 0.635            & 0.980             \\
Hamba \cite{dong2024hamba}       & \colorbox{yellow!30}{0.850}                   & \colorbox{yellow!30}{7.5}                     & 0.843                   & \colorbox{yellow!30}{7.7}                     & \colorbox{orange!30}{0.648}            & \colorbox{yellow!30}{0.982}             \\
WiLoR \cite{potamias2024wilor}        & \colorbox{orange!30}{0.851}                   & \colorbox{yellow!30}{7.5}                     & \colorbox{yellow!30}{0.846}                   & \colorbox{yellow!30}{7.7}                      & \colorbox{yellow!30}{0.646}            & \colorbox{orange!30}{0.983}             \\
\hline
HaMeR w/ {\datasetname}          &  \colorbox{orange!30}{0.851}                   & \colorbox{orange!30}{$7.4_{71}$}             & \colorbox{orange!30}{0.847}                   & \colorbox{orange!30}{$7.6_{76}$}           & 0.645            & \colorbox{red!30}{0.984}             \\
WiLoR w/ {\datasetname}          & \colorbox{red!30}{0.853}                   & \colorbox{red!30}{$7.3_{55}$}             & \colorbox{red!30}{0.848}                   & \colorbox{red!30}{$7.6_{24}$}           & \colorbox{red!30}{0.649}            & \colorbox{red!30}{0.984}             \\
\bottomrule
\end{tabular}}
\label{tab:suppl_rgb_ho3d}
\end{table*}

%% file: sec/5_RGBD_model.tex
\input{figs/fig_rgbd_flowchart}
\section{Assessing {\datasetname} on RGB-D settings}
\label{sec:method}

\input{tables/table_rgbd_ho3dv2}

\parhead{RGB-D architecture.}
Unlike the RGB-only setting, where we focus on evaluating the impact of \datasetname, the RGB-D setting requires an architectural change to fuse depth, as illustrated in Fig.~\ref{fig:rgbd_workflow}.
We use dual embedding branches to tokenize RGB and depth, followed by a lightweight bidirectional cross-attention module that exchanges information between the two modalities at corresponding image patches.
The fused tokens are then concatenated with task tokens and passed through the remaining transformer blocks for 3D hand pose estimation.
Note that,
the fusion module is designed to be lightweight and modular, allowing it to be easily integrated into existing ViT-based architectures like WiLoR~\cite{potamias2024wilor} with minimal changes. 

\parhead{Training.}
We conduct two training variants:
\emph{Real-only} trains on the real RGB-D datasets HO-3D~\cite{hampali2020ho3d} and DexYCB~\cite{chao2021dexycb}, while
\emph{Real + \datasetname} further trains together with our proposed {\datasetname}, following the same recipe and protocol as in the RGB experiments.

\parhead{Evaluation.}
We evaluate the models on HO-3D v2 \cite{hampali2020ho3d} and report on
(1) scale-translation aligned MPJPE (STA-MPJPE),
and (2) Procrustes-aligned MPJPE (PA-MPJPE),
where for both metrics, lower is better.

\parhead{Results.}
The quantitative results are reported in Tab.~\ref{tab:rgbd_ho3dv2}.
Our method consistently outperforms prior RGB-D approaches on both the STA-MPJPE (no rotation alignment) and PA-MPJPE (with rotation alignment) metrics, indicating that the improvement is not merely from more accurate global hand orientation,
but also a more accurate articulated hand structure after removing global similarity transforms.
Compared to Keypoint-Fusion~\cite{liu2024keypoint},
our model reduces the STA-MPJPE from 1.87\,cm to 1.09\,cm, a relative error reduction of approximately $41.7\%$.
Moreover, even the real-only variant remains competitive and already surpasses prior RGB-D baselines on STA and PA,
indicating that the fusion module is robust on its own, while mixed training with large-scale synthetic depth provides an additional boost.
As an ablation, removing RGB-D cross-attention leads to worse convergence and higher error, suggesting its effectiveness and importance in hinting the backbone to jointly consider RGB and depth cues on a given hand-like region.

\parhead{Estimation with missing depth.}
In real-world applications, depth maps are not always available.
We thus additionally evaluate our RGB-D model on HO-3D v2 by replacing the ground-truth depth maps with depth estimated from RGB using MoGe-v2~\cite{wang2025moge2accuratemonoculargeometry}.
Surprisingly,
this yields even better performance than using the ground-truth depth: STA-MPJPE improves from 1.09\,cm to 1.06\,cm, and PA-MPJPE improves from 0.81\,cm to 0.79\,cm.
This is likely because ground-truth HO-3D depth maps are heavily quantized and contain missing values, whereas MoGe-v2 produces smoother, denser estimated depths that may better match the synthetic training distribution, where background depth is also MoGe-based.

%% file: figs/fig_rgbd_flowchart.tex
\begin{figure*}[t]
    \centering
    \includegraphics[width=\linewidth]{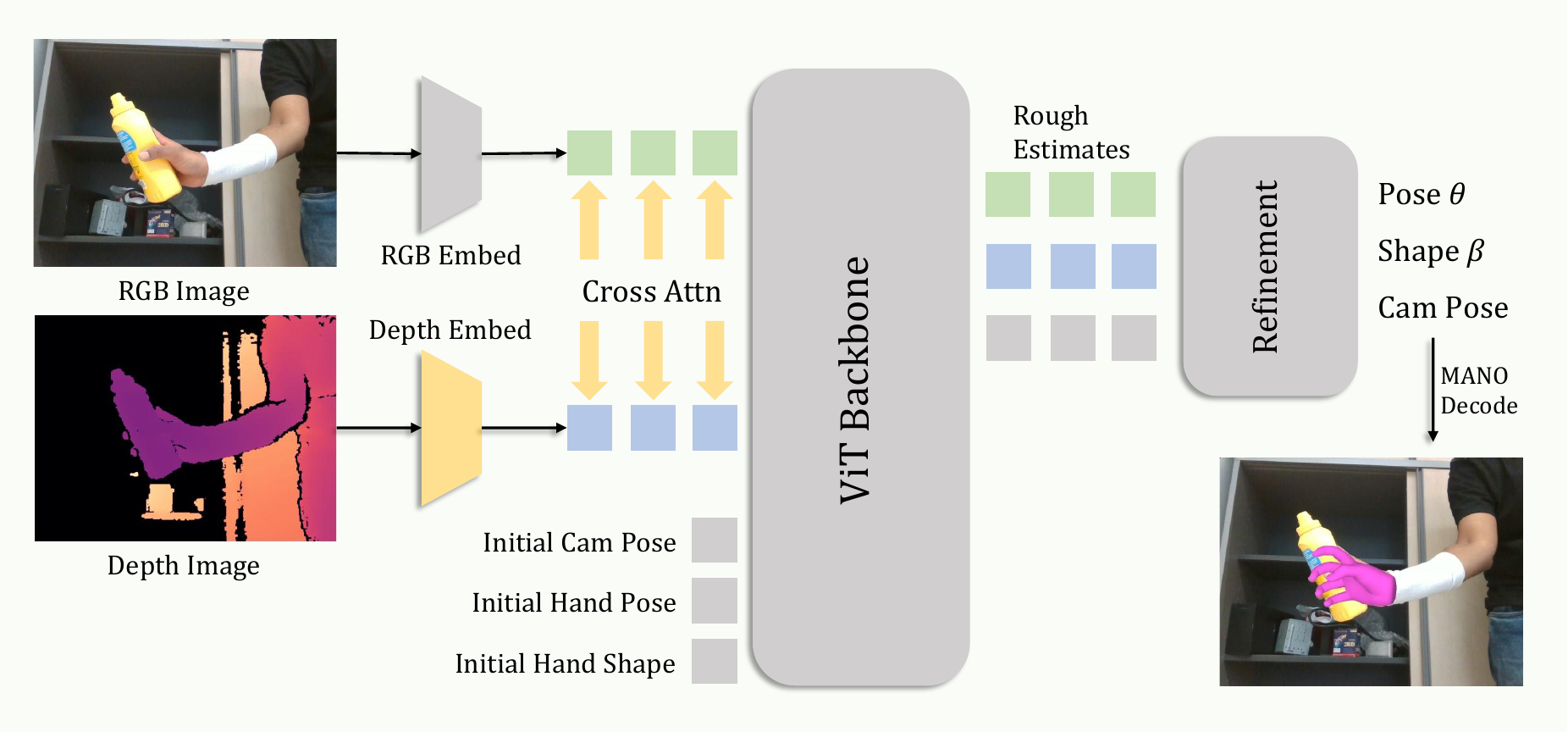}
    \caption{\textbf{Workflow of {\rgbdmodel}.}
    Built upon WiLoR~\cite{potamias2024wilor}’s RGB-only pipeline, we add a lightweight depth fusion module (highlighted in \colorbox{yellow!40}{yellow}).
    RGB and depth are embedded into parallel token sequences, followed by a bidirectional RGB-Depth cross-attention.
    The fused tokens are concatenated with other task tokens and fed into the ViT backbone and refinement head, whose outputs are finally 
    MANO decoded.}
    \label{fig:rgbd_workflow}
\end{figure*}

%% file: tables/table_rgbd_ho3dv2.tex
\begin{table*}[t]
\centering
    \caption{\textbf{Comparison with RGB-D methods on the \textbf{HO-3D v2} benchmark.}
    We report STA-MPJPE, and PA-MPJPE (in cm; $\downarrow$). 
    Top results are emphasized in \colorbox{red!30}{top1}, \colorbox{orange!30}{top2}, and \colorbox{yellow!30}{top3}.
    {\rgbdmodel} achieves the best overall results, and the real-only variant also surpasses prior RGB-D methods.
    Ablation on Xttn is also included. 
    \textbf{Note that baselines are trained in a dataset-specific setting, while \rgbdmodel targets cross-dataset generalization without benchmark-specific retraining.}
    }
\small
\begin{tabular}{llcc}
\toprule
\textbf{Input}  & \textbf{Method} & \textbf{STA-MPJPE} $\downarrow$ & \textbf{PA-MPJPE} $\downarrow$ \\
\midrule
RGB-D & DiffHand \cite{li2023diffhand}  & - & 0.93 \\
D\&PCL & IPNet \cite{ren2023ipnet}  & 2.01 & $0.87_{1}$ \\
RGB-D & Keypoint-Fusion \cite{liu2024keypoint}  & 1.87 & $0.94_3$  \\
\midrule
RGB-D & {\rgbdmodel} (Real-only)  & $1.20_1$ & $0.89_1$ \\
& Eval w/ MoGe-2~\cite{wang2025moge2accuratemonoculargeometry} Estimated Depth  & \colorbox{yellow!30}{$1.18_{4}$} & \colorbox{orange!30}{$0.80_{2}$} \\
RGB-D & {\rgbdmodel} (Real + {\datasetname}) & \colorbox{orange!30}{$1.09_{7}$} & \colorbox{yellow!30}{$0.81_{4}$} \\
& Eval w/ MoGe-2~\cite{wang2025moge2accuratemonoculargeometry} Estimated Depth & \colorbox{red!30}{$1.06_{8}$} & \colorbox{red!30}{$0.79_{3}$} \\
& w/o RGB-D Cross Attention & {$1.21_1$} & {$0.85_5$} \\
\bottomrule
\end{tabular}
    \label{tab:rgbd_ho3dv2}
\end{table*}

%% file: sup_sec/dataset.tex
\section{\datasetname dataset details}
\label{sec:suppl_dataset}

\subsection{Synthetic data generation pipeline}
\label{subsec:sup_data_gen}

We provide additional implementation details of \datasetname generation below.

\parhead{FOV range.}
For each rendered view, the camera field of view is sampled uniformly from $30^\circ$ to $40^\circ$.

\parhead{Camera distance distribution.}
We sample the hand--camera distance from Gaussian distributions with means of $0.6$m, $0.7$m, and $1.0$m, each with a standard deviation of $0.1$m. This produces both close-up and distant views while keeping the hand at a realistic scale in the image.

\parhead{Viewpoint sampling strategy.}
During rendering, the MANO hand is placed at the origin
and a forearm mesh is attached at the wrist. We then sample a camera center at the chosen distance along a random 3D viewing direction, and orient the camera to look at the origin. This allows the hand--arm pair to be observed from diverse viewpoints in 3D space.

\parhead{Lighting configuration.}
We use at most five lights per scene.
For each scene, we randomly choose the number of lights from one to five.
The ambient illumination is first set to roughly match the dominant color tone of the sampled background, as described in the main text, to maintain visual consistency between the rendered foreground and background.  
The remaining lights are randomly chosen from three types: \texttt{point}, \texttt{directional}, and \texttt{spot}.
For each light, we randomize its placement and associated parameters, and assign either a vivid color with some probability or a near-white to slightly warm tone otherwise.
We also randomly enable or disable shadows and vary the corresponding shadow ranges.
This strategy increases illumination diversity while preserving overall scene coherence.

\parhead{Processing GraspXL sequences.}
GraspXL~\cite{zhang2024graspxl} provides hand-object interaction sequences, where the hand approaches an object, establishes contact, and then releases it. 
For each sample in {\datasetnamegxl}, we first randomly select an object and then randomly choose one interaction sequence associated with that object. 
Since our goal is to capture occluded hands during active manipulation, we select the exact frame where the hand is grabbing on the object. 
This frame is then used for rendering the hand-object sample.

The original GraspXL sequences are built on simplified, texture-free object meshes for efficient physics simulation. 
To improve visual realism, we retrieve the corresponding textured object meshes from Objaverse~\cite{deitke2023objaverse} and align them to the simplified meshes used in GraspXL. 
Specifically, we scale the textured mesh to match the size of the simulation mesh and use it during rendering, while preserving the hand pose and object placement from the selected GraspXL frame. 
The remaining rendering process, including camera sampling, lighting, background composition, and depth generation, follows the same pipeline as the single-hand branch.

\input{tables/table_suppl_dataset_details}

In summary, the details of \datasetname are listed in Tab.~\ref{tab:suppl_anyhand_summary}.

\subsection{Additional visualization of \datasetname}

We provide additional qualitative examples from \datasetname in Fig.~\ref{fig:supple_more_anyhand_vis}. 
For each sample, we show the rendered RGB image together with its corresponding 3D hand mesh, 2D joint annotations, and depth map. 
These examples illustrate the diversity of \datasetname in hand pose, camera viewpoint, appearance, and interaction setting, while highlighting its rich multi-modal annotations.

\input{figs/fig_suppl_more_AnyHand_vis}

%% file: tables/table_suppl_dataset_details.tex
\begin{table*}[tb!]
\centering
\small
\setlength{\tabcolsep}{5pt}

\caption{\textbf{Summary of the \datasetname generation pipeline and dataset statistics.}
The table summarizes the main design choices, rendering settings, and annotations used in \datasetname.}
\label{tab:suppl_anyhand_summary}

\begin{tabular}{ll}
\toprule
\textbf{Component} & \textbf{Summary} \\
\midrule
Hand shapes & 47,438, sampled from real datasets \\
Hand poses & On-the-fly samples from DPoser-Hand~\cite{lu2025dposerx} \\
Hand textures & 10,240 unique textures \\
Arm textures & 254 textures \\
Backgrounds & 1,270 indoor images and HDRI maps \\
Objects & $>$500K objects \\
Camera FOV & Randomly sampled from $30^\circ$ to $40^\circ$ \\
Camera distance & Mean $0.6$m / $0.7$m / $1.0$m, std $0.1$m \\
Lights & 1 to 5 random lights per scene \\
Views per scene & 2 \\
AnyHand-Single & 1.25M scenes, 2.5M images \\
AnyHand-Interact & 2.05M scenes, 4.1M images \\
Stored annotations & RGB, depth, mask, bbox, camera, 3D pose/shape \\
\bottomrule
\end{tabular}
\end{table*}

%% file: figs/fig_suppl_more_AnyHand_vis.tex
\begin{figure*}[tbh!]
    \centering
    \includegraphics[width=0.8\linewidth]{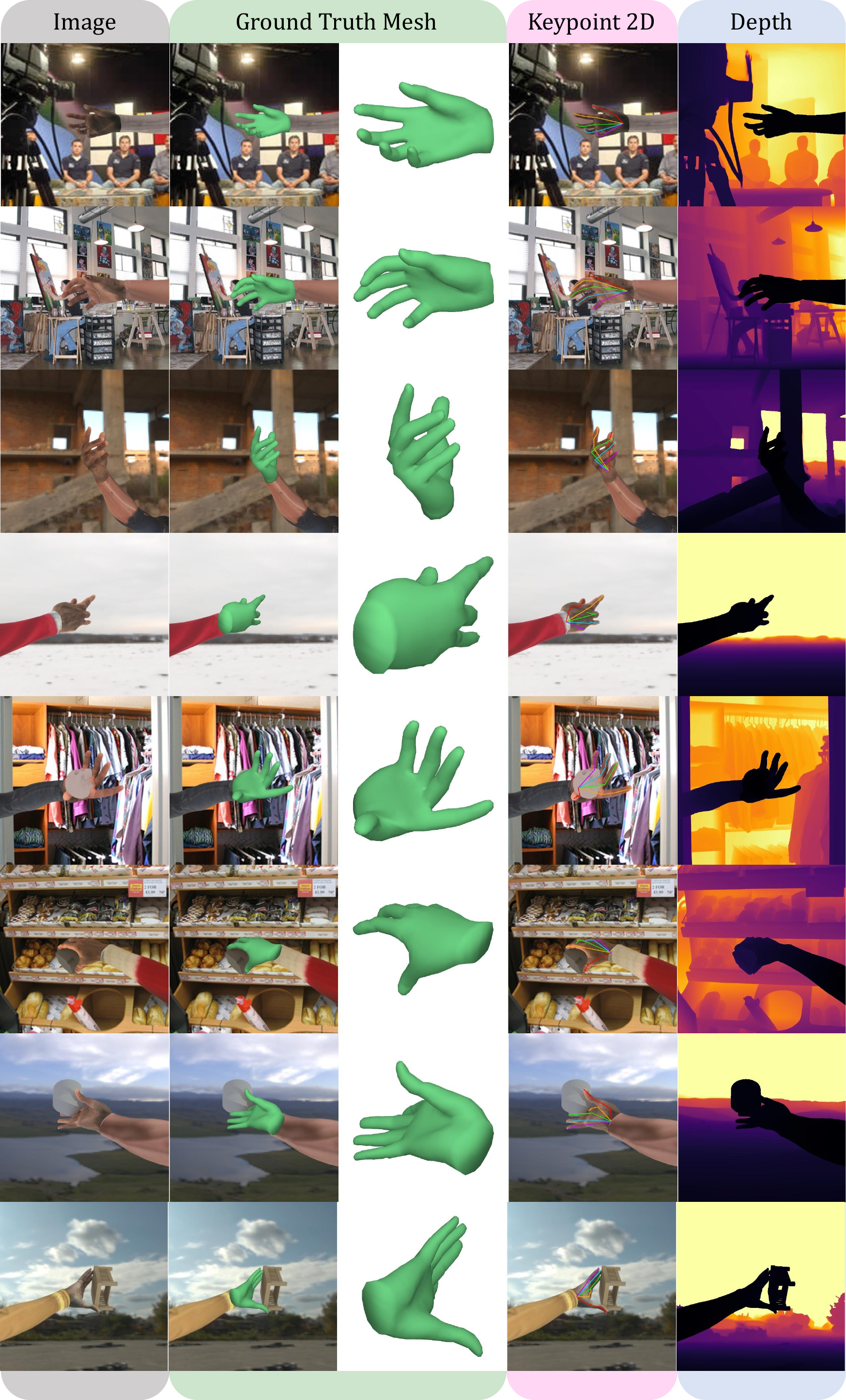}
    \caption{\textbf{Qualitative visualizations of \datasetname.} Additional examples from \datasetname, showing the rendered RGB images together with their corresponding 3D hand meshes, 2D joint annotations, and depth maps. 
    These examples illustrate the diversity of poses, appearances, viewpoints, and interaction scenarios covered by the dataset.}
    \label{fig:supple_more_anyhand_vis}
\end{figure*}

%% file: sup_sec/related_extended.tex
\section{Additional discussion on synthetic hand datasets}
\label{app:synthetic_dataset_discussion}

\parhead{Overview.}
Tab.~\ref{tab:synthetic_dataset_summary} summarizes representative synthetic datasets for 3D hand pose estimation.
Here, we provide a more detailed comparison along several data-design axes: pose source, hand-shape and appearance diversity, object coverage, rendering variation, arm context, and depth supervision.
This axis-based comparison is important because prior datasets often improve one specific aspect of synthetic hand data, but do not jointly provide million-scale RGB-D supervision, hand-object occlusion, explicit arm context, broad appearance variation, and diverse 3D annotations.

\parhead{Early synthetic RGB-D supervision.}
RHD~\cite{zimmermann2017rhd} is one of the earliest synthetic resources for RGB-based 3D hand pose estimation.
It is built from 20 animated characters performing 39 actions, with a split over characters and actions to reduce train-test overlap.
It provides RGB images, segmentation masks, 21-keypoint 2D/3D annotations, camera intrinsics, and depth maps, making it an important early reference for aligned synthetic supervision.
Its main limitation is not annotation completeness, but scale and coverage: the dataset contains only 44k images and focuses on isolated hands without object-induced occlusion or explicit hand--forearm continuity.
Therefore, RHD establishes the feasibility of synthetic RGB-D supervision, but it is not designed for the current data-scaling regime of foundation-style hand reconstruction.

\parhead{Synthetic hand-object interaction.}
ObMan~\cite{hasson2019obman} introduced synthetic hand-object manipulation with 3D hand and object meshes, segmentation masks, and depth maps, and improves realism by rendering embodied hands rather than floating-hand synthesis.
However, its diversity and scale remain limited by modern standards.
ObMan contains 2,772 ShapeNet objects from 8 everyday categories and 21K grasp configurations, resulting in a relatively constrained space of object shapes, hand poses, and interaction patterns.
Its hand appearance is based on 176 high-resolution hand textures from 20 subjects, and its rendering relies on LSUN/ImageNet background images and Blender compositing rather than diverse HDR environment illumination.
In contrast, \datasetname-Interact provides million-scale hand-object supervision with broader pose, texture, object, occlusion, and lighting diversity.



\parhead{Two-hand interaction data.}
Re:InterHand~\cite{moon2023re-interhand} focuses on reducing the appearance gap in two-hand interaction data.
It captures interacting-hand motions from 10 subjects in a multi-camera studio with 170 synchronized cameras and 469 lights, and relights the tracked hands using 2,144 high-resolution environment maps.
Its data therefore provides realistic two-hand poses and relighted appearances, but its scope is limited to hand--hand interaction.
It does not include manipulated objects, object-induced occlusions, aligned depth maps, or arm context around the wrist.
Its subject-level shape diversity is also bounded by the 10 captured subjects.
Therefore, Re:InterHand addresses the realism of two-hand images, but does not target scalable hand-object reconstruction or RGB-D hand pose learning.

RenderIH~\cite{li2023renderih} also targets two-hand interaction data.
It starts from InterHand2.6M poses and augments them with random joint offsets to increase pose variation and reduce unrealistic hand--hand interpenetration.
It renders 1.2M images with varied viewpoints, 300 HDR panoramas, and 30 base hand textures with additional shader-level skin-tone perturbations.
However, because its pose generation is based on InterHand2.6M, its pose distribution remains closely tied to the original InterHand pose space.
{In addition, RenderIH does not model arm or wrist context: the hand meshes terminate at an open wrist boundary.
This omission may affect training quality, as Zhao \etal~\cite{zhao2025analyzing} show that arm context is a considerable factor in reducing the synthetic-to-real gap for hand pose estimation.
Therefore, while RenderIH improves the visual diversity of two-hand renderings, it does not address wrist-boundary artifacts or provide the broader arm context needed for robust real-image generalization.}
Thus, RenderIH is better viewed as an augmentation and rendering pipeline for two-hand interaction poses, rather than a newly generated large-scale hand dataset with a substantially different pose source.

In contrast, \datasetname targets a different setting.
It provides scalable RGB-D supervision for both single-hand and hand-object scenes, with jointly rendered hand, forearm, object, and depth annotations.
Compared with Re:InterHand and RenderIH, \datasetname explicitly covers hand-object manipulation, object-induced occlusion, aligned depth supervision, and arm context, while also using a larger appearance source with 10,240 diffused hand appearances and 254 forearm textures.

\parhead{Relation to synthetic-to-real gap analysis.}
A closely related recent work is Zhao \etal~\cite{zhao2025analyzing}, which studies the synthetic-to-real gap in 3D hand pose estimation using benchmark-matched synthetic counterparts.
Unlike the datasets discussed above, their work is primarily a controlled analysis framework rather than a large-scale training resource.
They explicitly decompose factors such as hand texture, background, arm context, object occlusion, pose distribution, and skeleton topology.
This objective is different from ours: Zhao \etal aim to minimize confounding factors and stay close to target benchmark distributions, whereas {\datasetname} aims to build a scalable synthetic data-generation pipeline that extends beyond the coverage of existing real datasets.

The difference is reflected in pose, appearance, and scale.
Zhao \etal align their synthetic pose distribution with target real datasets such as FreiHAND and Dex-YCB by fitting NIMBLE to MANO annotations.
Therefore, their synthetic data is largely derived from existing benchmark poses and can be viewed as benchmark-conditioned rendering and augmentation, rather than a new dataset with an independent pose source.
They use NIMBLE-based hand textures generated from 38 photo-realistic hand texture assets and 669 HDRI scenes.
By contrast, {\datasetname} samples poses on the fly from the DPoser-Hand diffusion prior trained on multiple real datasets, and uses Handy-based texture generation to obtain 10,240 unique hand appearances.
In addition, {\datasetname} attaches textured forearm geometry using 254 SMPLitex body textures and renders hand--forearm geometry jointly in simulation, which avoids the boundary artifacts that can arise from compositional arm pasting.

\parhead{Takeaway.}
Overall, prior synthetic hand datasets have made important progress along different axes:
RHD establishes early aligned RGB-D supervision;
ObMan introduces synthetic hand-object interaction with depth and segmentation;
Re:InterHand improves realism for relighted two-hand interaction;
RenderIH improves synthetic two-hand pose optimization and photorealistic rendering;
and Zhao \etal provide a controlled analysis of the synthetic-to-real gap.
{\datasetname} is designed to address a complementary but increasingly important gap:
a simulation-native, million-scale RGB-D training resource that jointly covers diverse hand poses, large hand-shape and texture variation, explicit forearm context, large-scale object-induced occlusion, realistic rendering variation, and aligned geometric annotations.

%% file: sup_sec/computation.tex
\section{Compute resources}
\label{app:detail}
Our main training experiments were conducted on a GPU cluster. 
Unless otherwise specified, each main run used a single node with 8 NVIDIA H100 GPUs, 128 CPU threads from dual Intel Xeon Platinum 8350C CPUs, and approximately 1\,TiB of system memory. 
Each full training run took approximately 5--7 days under this setup. 
We used the same evaluation pipeline for all reported checkpoints to ensure consistent comparison across baselines and variants.

%% file: sup_sec/limitation.tex
\section{Limitations}
\label{sec:sup_limitations}

While \datasetname provides large-scale and diverse RGB-D supervision for hand pose estimation, it also has several limitations that suggest useful directions for future work.

First, hand poses are sampled from DPoser-Hand~\cite{lu2025dposerx}, a learned diffusion prior trained on a finite collection of real datasets. Although we apply biomechanical filtering (BMC loss ~\cite{yang2021cpf}) to remove implausible poses, some anatomically unlikely configurations may still pass through the filter.
Future work may further improve pose sampling by incorporating stronger anatomical constraints or by adapting the pose prior to specific downstream domains.

Second, our composition pipeline estimates background depth and composites the rendered hand onto 2D background images, rather than placing the hand inside a full 3D scene. This prioritizes scalability but means the background depth near hand boundaries may not be perfectly consistent.
In practice, this is acceptable 
because most hand pose estimation pipelines operate on tightly cropped hand regions, 
where the learning signal is dominated by the hand itself. 
Rendering complete 3D scenes at this scale would require substantially higher asset and computation costs
, which would reduce the efficiency of large-scale dataset construction. 
A useful future direction is to develop an efficient scene-level rendering pipeline for large synthetic hand datasets.

Finally, \datasetname is built by integrating existing tools for pose sampling, texture generation, and rendering. Its primary contribution is as a large-scale data resource for the community. Beyond the dataset itself, we provide systematic studies on data scaling, real-synthetic mixing, and component design that can inform future synthetic data efforts for hand pose estimation. Leveraging \datasetname to train a unified foundation model for both RGB and RGB-D hand pose estimation is a promising direction that we leave for future exploration.

%% file: sup_sec/benchmark.tex
\section{Benchmark on \datasetname test set}
\label{sec:suppl_benchmark}

\input{tables/table_rgb_benchmark}

We evaluate state-of-the-art RGB methods, including HaMeR~\cite{pavlakos2024hamer} and WiLoR~\cite{potamias2024wilor}, as well as their variants trained with {\datasetname} (as mentioned in the RGB experiments), on {\datasetname} as a benchmark. 

As shown in Tab.~\ref{tab:rgb_benchmark}, 
the original HaMeR and WiLoR models exhibit non-trivial generalization to \datasetname, despite not being trained on this data distribution. 
However, adding \datasetname in training consistently improves performance
across all four splits and across both pose and mesh metrics, suggesting that exposure to \datasetname during training substantially reduces the domain gap
to the \datasetname test set.

The gains are particularly pronounced on the hand-object interaction splits (\datasetnamegxl). 
On \datasetnamegxl-EnvMap and \datasetnamegxl-Indoor, 
adding \datasetname in training reduces MPJPE from roughly 19--26\,mm to about 5--6\,mm, while raising F@5 from around 0.09--0.11 to about 0.56--0.61. These improvements correspond to large gains in both pose accuracy and mesh reconstruction quality under challenging interaction scenarios.
On the single-hand splits (\datasetnamedpx), the improvements are smaller but still consistent across metrics, demonstrating that the benefits of \datasetname extend beyond interaction-heavy cases. 
Overall, these results support the use of \datasetname both as a scalable training source and as a reference benchmark for future methods.

%% file: tables/table_rgb_benchmark.tex

\begin{table*}[t!]
\centering
\caption{\textbf{Benchmarking methods on \datasetname}. 
Best results in each sub-table are shown in \textbf{bold}. 
For F-scores, we report both pre-alignment and post-alignment values. 
These results serve as reference values for future methods utilizing \datasetname.}
\label{tab:rgb_benchmark}
\resizebox{\textwidth}{!}{%
\begin{tabular}{@{}l cccccc cccc@{}}
\toprule
& \multicolumn{6}{c}{\textbf{Pose Metrics}} & \multicolumn{4}{c}{\textbf{Mesh Metrics}} \\
\cmidrule(lr){2-7} \cmidrule(lr){8-11}
Method & AUC$_j$ $\uparrow$ & MPJPE $\downarrow$ & AUC$_{j\_pa}$ $\uparrow$ & PA-MPJPE $\downarrow$ & AUC$_{j\_sta}$ $\uparrow$ & STA-MPJPE $\downarrow$ & F@5 $\uparrow$ & F@15 $\uparrow$ & F-al@5 $\uparrow$ & F-al@15 $\uparrow$ \\
\midrule
\multicolumn{11}{c}{{{\datasetnamedpx}-EnvMap}} \\
\midrule
HaMeR              & 0.600 & 21.290 & 0.800 & 10.090 & 0.624 & 19.840 & 0.098 & 0.552 & 0.293 & 0.854 \\
WiLoR              & 0.567 & 23.486 & 0.778 & 11.199 & 0.588 & 22.088 & 0.090 & 0.518 & 0.245 & 0.817 \\
HaMeR w/ \datasetname & 0.701 & 14.980 & 0.890 &  5.500 & 0.805 &  9.780 & 0.139 & 0.625 & 0.593 & 0.967 \\
WiLoR w/ \datasetname & \textbf{0.715} & \textbf{14.290} & \textbf{0.904} & \textbf{4.820} & \textbf{0.811} & \textbf{9.530} & \textbf{0.140} & \textbf{0.658} & \textbf{0.681} & \textbf{0.973} \\
\midrule
\multicolumn{11}{c}{{{\datasetnamedpx}-Indoor}} \\
\midrule
HaMeR              & 0.573 & 23.020 & 0.788 & 10.690 & 0.596 & 21.460 & 0.093 & 0.523 & 0.265 & 0.836 \\
WiLoR              & 0.542 & 25.355 & 0.763 & 11.974 & 0.557 & 24.171 & 0.086 & 0.496 & 0.217 & 0.791 \\
HaMeR w/ \datasetname & \textbf{0.755} & \textbf{12.400} & \textbf{0.857} & \textbf{7.160} & \textbf{0.792} & \textbf{10.560} & \textbf{0.191} & \textbf{0.744} & \textbf{0.455} & \textbf{0.929} \\
WiLoR w/ \datasetname & 0.731 & 13.780 & 0.821 & 9.020 & 0.752 & 12.740 & 0.182 & 0.712 & 0.346 & 0.881 \\
\midrule
\multicolumn{11}{c}{{{\datasetnamegxl}-EnvMap}} \\
\midrule
HaMeR              & 0.638 & 19.250 & 0.801 & 10.030 & 0.664 & 17.860 & 0.107 & 0.582 & 0.260 & 0.840 \\
WiLoR              & 0.579 & 23.304 & 0.772 & 11.523 & 0.603 & 21.946 & 0.091 & 0.507 & 0.200 & 0.789 \\
HaMeR w/ \datasetname & \textbf{0.891} &  \textbf{5.480} & 0.933 &  3.360 & \textbf{0.892} &  5.460 & 0.601 & \textbf{0.956} & 0.827 & \textbf{0.986} \\
WiLoR w/ \datasetname & 0.890 &  5.510 & \textbf{0.934} &  \textbf{3.300} & \textbf{0.892} &  \textbf{5.440} & \textbf{0.607} & 0.951 & \textbf{0.829} & \textbf{0.986} \\
\midrule
\multicolumn{11}{c}{{{\datasetnamegxl}-Indoor}} \\
\midrule
HaMeR              & 0.595 & 21.990 & 0.795 & 10.360 & 0.642 & 19.380 & 0.096 & 0.526 & 0.247 & 0.828 \\
WiLoR              & 0.545 & 25.594 & 0.768 & 11.757 & 0.583 & 23.332 & 0.085 & 0.464 & 0.195 & 0.778 \\
HaMeR w/ \datasetname & \textbf{0.887} &  \textbf{5.690} & \textbf{0.932} &  \textbf{3.400} & \textbf{0.889} &  \textbf{5.590} & \textbf{0.579} & \textbf{0.952} & \textbf{0.883} & \textbf{0.986} \\
WiLoR w/ \datasetname & 0.882 &  5.940 & 0.931 &  3.440 & 0.885 &  5.770 & 0.560 & 0.944 & 0.817 & 0.985 \\
\bottomrule
\end{tabular}%
}
\end{table*}